\documentclass{article}

\usepackage[numbers,sort&compress]{natbib}

\usepackage{microtype}
\usepackage{graphicx}
\usepackage{subfigure}
\usepackage{booktabs} % for professional tables
\usepackage{textcomp}

\usepackage[final]{neurips_2021}
\usepackage{color}
\usepackage{multirow}
\usepackage{bbm}
\usepackage{bm}
\usepackage{amsfonts,amssymb}
\usepackage{amsmath}
\usepackage{graphicx} %use graph format
\usepackage{epstopdf}

\usepackage[all]{xy}
\usepackage{amsmath}
\usepackage{wrapfig}
\newcommand{\tabincell}[2]{\begin{tabular}{@{}#1@{}}#2\end{tabular}}

\usepackage{algorithmic}
\usepackage{algorithm}
\usepackage{amsmath}
\usepackage[utf8]{inputenc} % allow utf-8 input
\usepackage[T1]{fontenc}    % use 8-bit T1 fonts
\usepackage{hyperref}       % hyperlinks
\usepackage{url}            % simple URL typesetting
\usepackage{booktabs}       % professional-quality tables
\usepackage{amsfonts}       % blackboard math symbols
\usepackage{nicefrac}       % compact symbols for 1/2, etc.
\usepackage{microtype}      % microtypography
\usepackage{xcolor}         % colors
\usepackage{stfloats}

\newcommand{\Wmat}[0]{\ensuremath{{\bf W}} }

\newcommand{\bv}[0]{\ensuremath{\boldsymbol{b}} }

\newcommand{\hv}[0]{\ensuremath{\boldsymbol{h}} }

\newcommand{\kv}[0]{\ensuremath{\boldsymbol{k}} }

\newcommand{\xv}[0]{\ensuremath{\boldsymbol{x}} }
\newcommand{\yv}[0]{\ensuremath{\boldsymbol{y}} }

\newcommand{\Ev}[0]{\ensuremath{\boldsymbol{E}} }

\newcommand{\Lv}[0]{\ensuremath{\boldsymbol{L}} }

\newcommand{\Nv}[0]{\ensuremath{\boldsymbol{N}} }

\newcommand{\lambdav}[0]{\ensuremath{\boldsymbol{\lambda}} }

\newcommand{\Sigmamat}[0]{\ensuremath{\boldsymbol{\Sigma}} }

\newcommand{\muv}[0]{\ensuremath{\boldsymbol{\mu}} }

\newcommand{\given}{\,|\,}

\title{TopicNet: Semantic Graph-Guided Topic Discovery}
\author{%
  Zhibin Duan, Yishi Xu, Bo Chen\thanks{ Corresponding Author.} ,
  Dongsheng Wang, Chaojie Wang \\
  National Laboratory of Radar Signal Processing, Xidian University, Xi’an, China\\
  \texttt{xd\_zhibin@163.com, bchen@mail.xidian.edu.cn} 
  \And
   Mingyuan Zhou\\
   McCombs School of Business, The University of Texas at Austin\\
  \texttt{mingyuan.zhou@mccombs.utexas.edu}

}

\begin{document}

\maketitle

\begin{abstract}
Existing deep hierarchical topic models are able to extract semantically meaningful topics from a text corpus  in an unsupervised manner and automatically organize them into a topic hierarchy. However, it is unclear how to incorporate prior belief such as knowledge graph to guide the learning of the topic hierarchy. To address this issue, we introduce TopicNet as a deep hierarchical topic model that can inject prior structural knowledge as an inductive bias to influence the learning. TopicNet represents each topic as a Gaussian-distributed embedding vector, projects the topics of all layers into a shared embedding space, and explores both the symmetric and asymmetric similarities between Gaussian embedding vectors to incorporate prior semantic hierarchies. With an auto-encoding variational inference network, the model parameters are optimized by minimizing the evidence lower bound and a regularization term via stochastic gradient descent. Experiments on widely used benchmarks show that TopicNet outperforms related deep topic models on discovering deeper interpretable topics and mining better document~representations.
\end{abstract}
\section{Introduction} \label{sec:introduction}
\begin{figure}
%\vspace{-20pt}
\begin{center}
\includegraphics[width=.9\textwidth]{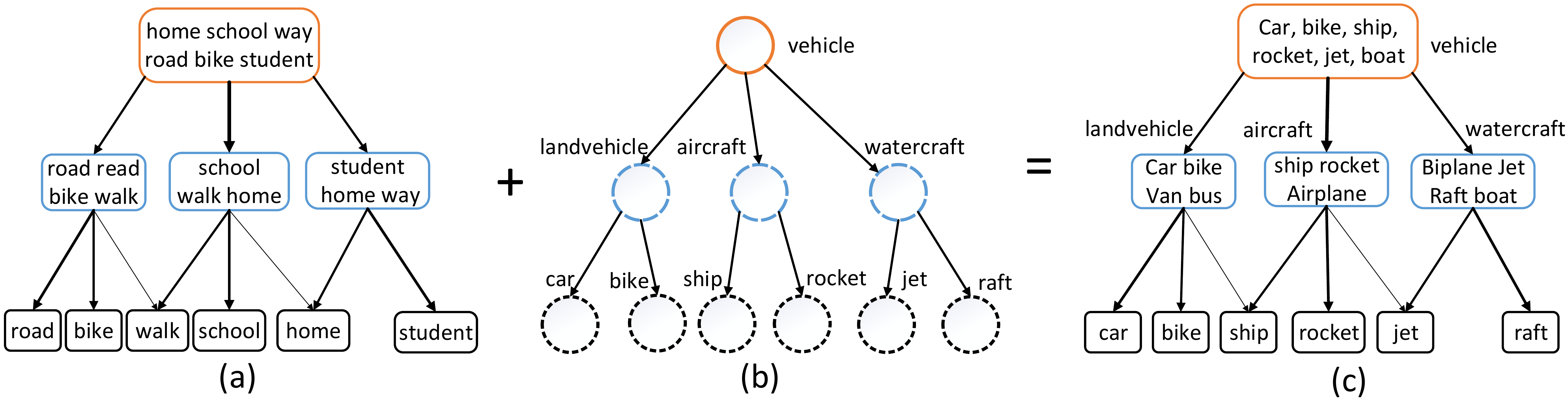}
\end{center}
\vspace{-12pt}
\caption{\small Illustration of (a) topics  discovered by a hierarchical topic model, (b) semantic graph constructed by prior knowledge, and (c) topics learned by TopicNet, the proposed knowledge-based hierarchical topic model.}
\label{fig:motivation}
\vspace{-14pt}
\end{figure}

Topic models, which have the ability to uncover the hidden semantic structure in a text corpus, have been widely applied to text analysis. %topic modeling has been a successful technique for text analysis for almost two decades. 
Generally, a topic model is designed to discover a set of semantically-meaningful latent topics from a collection of documents, each of which captures word co-occurrence patterns commonly observed in a document. %describes an interpretable semantic concept. 
While vanilla topic models, such as latent Dirichlet allocation (LDA) \cite{blei2003latent} and Poisson factor analysis (PFA) \cite{zhou2012beta}, are able to achieve this goal, a series of their hierarchical extensions \cite{blei2010nested, paisley2014nested, gan2015scalable, zhou2016augmentable, cong2017deep, zhao2018Dirichlet,guo2018deep,wang2018multimodal,wang2019convolutional,wang2020deep} have been developed in the hope of exploring multi-layer document representations and mining meaningful topic taxonomies. 
Commonly, these hierarchical topic models aim to learn a hierarchy, in which the latent topics exhibit an increasing level of abstraction when moving towards a deeper hidden layer, as shown in Fig.~\ref{fig:motivation}(a). 
Consequently, it provides users with an intuitive and interpretable way to better understand textual data.
%To discover the underlying semantic structure from raw data, there is a surge of interest in hierarchical probabilistic generative model  \cite{blei2003latent,hinton2006fast,lee2009convolutional,chen2013deep,zhou2016augmentable,wang2019convolutional}, which assume there is a hierarchical structure where certain aspects are abstractions of others. Especially, topic modeling has enjoyed great success in semantic structure discovery for text data. In general, topic models like Latent Dirichlet Allocation \cite{blei2003latent} and Poisson factor analysis  \cite{zhou2012beta}, aim to discover a set of latent topics from a collection of documents, each of which describes an interpretable semantic concept. As the extensions of single layer topic models, hierarchical topic models such as  \cite{blei2010nested, paisley2014nested, gan2015scalable, zhou2016augmentable, zhang2018whai, duan2021sawtooth}, has enjoyed great success in hierarchical semantic structure discovering and its applications in different tasks \cite{wang2018multimodal,Wang2020Learn, Zhang2020Variational, wang2020deep,guo2020recurrent}. All of these models assuming the higher layer concept is the mixture of the lower layer concepts. and as shown in Fig.~\ref{fig:motivation} (a), the latent topics exhibit an increasing level of abstraction when moving towards a deeper hidden layer. 

Despite their attractive performance,
%and recent popularity, 
many existing hierarchical topic models are purely data-driven and  incorporate no prior domain knowledge, which may result in some learned topics failing to describe a semantically coherent concept \cite{chang2009reading}, especially for those at a deeper hidden layer \cite{duan2021sawtooth}. %{\color{blue}
Furthermore, the inflexibility of adding prior knowledge also somewhat limits the applicability of hierarchical topic models, since %cases are 
it is common that a user is interested in a specific topic structure and only focused on information related to it \cite{meng2020hierarchical}.
%}. 
To address this issue, %overcome this shortcoming,
%we consider explicitly introducing prior knowledge into the process of topic modeling. Specifically, 
we assume a structured prior in the form of a predefined topic hierarchy, where each topic is described by a semantic concept, with the concepts at adjacent layers following the hypernym relations, as illustrated in Fig.~\ref{fig:motivation}(b). Such a hierarchy can be easily constructed, either  by some generic knowledge graph like WordNet \cite{miller1995wordnet} or according to the user's customization. {%\color{blue}
However, there are two main challenges:
%are mainly considered: 
one is how to model this semantic hierarchy, and the other is how to combine it with topic models.} 
%Though achieving appealing performance, most of the existing hierarchical topic models are data-driven models, which do not consider the prior knowledge from the real world. Corresponding to hierarchical semantic structure of topic models, there is the hypernym relation between concepts in the real world. For example, a hypernym pair in the WordNet \cite{miller1995wordnet} is a pair of concepts where the first concept is a specialization or an instance of the second, e.g., (car, lanvehivle) or (landvehicle, vehicles). And shown in Fig.~\ref{fig:motivation} (b),  a general concept such as ``vehicles'' is an abstraction of more specific concepts such as ``landvehicle'' or ``aircraft''. These concepts can constitutes a hierarchical semantic structure by hypernym relation between themself. From Fig.~\ref{fig:motivation} (a) and Fig.~\ref{fig:motivation}(b), we can see that WordNet have similar assumption with topic models. With this observation, we consider how incorporate these hypernym relation between concepts to topic models remains an open and important research problem. The expected result shown in Fig.~\ref{fig:motivation} (c), which combine topic model with TopicTree, and we can know the meaning of each topics. 

For the first challenge, a general and powerful approach \cite{chopra2005learning} is to learn a distance-preserving mapping, which maps semantically similar topics to nearby points in the embedding space, with a symmetric distance measure ($e.g.$, Euclidean or cosine distance) typically being used. However, the embeddings learned by such a scheme can not perfectly reflect the entailment between concepts. To mitigate this issue,  \citet{vendrov2015order} exploit the partial order structure of the semantic hierarchy %and propose a method 
to learn order embeddings that respect this partial order structure. On the other hand, considering that probability distributions are better at capturing uncertainties of concepts than point vectors, \citet{athiwaratkun2018hierarchical} introduce density order embeddings as an improvement to order embeddings. Density order embeddings allows the entailment relationship to be expressed naturally, with general words such as “entity” corresponding to broad distributions that encompass more specific words such as “animal.” Through encapsulation of probability densities, it can intuitively reflect the hierarchical structure of semantic concepts, thus offering us an effective way to model the topic hierarchy.
%To capture the visual-semantic hierarchy structure, a poplar method is to project the structure into the vector space.  \cite{vendrov2015order} propose order-embeddings, which exploit the partial order structure of the visual-semantic hierarchy by learning a mapping which is not distance-preserving but order-preserving between the visual-semantic hierarchy and a partial order over the embedding space. And  \cite{athiwaratkun2018hierarchical} introduce density order embeddings, which learn hierarchical representations through encapsulation of probability densities. It is a challenging task that combining the above methods with topic models not long ago. 

As for the second challenge of combining the constructed topic hierarchy with topic models, there are two major difficulties. For one thing, the structured prior requires modeling the relationship between topics across different layers directly, while hierarchical topic models often assume that the topics at different layers are independently drawn in the prior \cite{zhou2016augmentable}, which makes the fusion of the two a bit far-fetched. For another, since the constructed topic hierarchy can be very deep for large-scale text corpora, an equally deep topic model with multiple latent layers is needed to incorporate with it. However, most existing deep neural topic models are hard to go very deep due to the problem of posterior collapse \cite{2016Variational,2016A,2017Neural,2017Towards,sonderby2016ladder, zhang2018whai, duan2021sawtooth}. Fortunately, SawETM \cite{duan2021sawtooth}, a recently proposed neural topic model, not only
builds the dependency between topics at different layers in a shared embedding space, but also alleviates the issue of posterior collapse to some extent through a novel network module referred to as Sawtooth Connection (SC), 
and hence overcoming both difficulties. 

In this paper, a novel knowledge-based hierarchical topic model has been proposed, the core idea of which is to represent each topic as a Gaussian embedding, and then project the topics of all layers into a shared embedding space to facilitate the injection of structured prior knowledge into the model. In particular, we first introduce the Gaussian-distributed embeddings to the SawETM, resulting in a variant of SawETM with a stochastic decoder called Gaussian SawETM. To incorporate prior belief into Gaussian SawETM, we then %enforce constraints on 
constrain the topics at adjacent layers to encourage them to capture the concepts satisfying the predefined hypernym relations. With auto-encoding variational inference, the entire model is learned in an end-to-end manner by minimizing the evidence lower bound and a regularization term. Extensive experiments demonstrate our model has competitive performance and better interpretability in comparison to most existing hierarchical topic models.

The main contributions of the paper can be summarized as follows:
\begin{itemize}
\item To capture semantic uncertainties of words and topics, we propose a novel probabilistic generative model with a stochastic decoder, referred to as Sawtooth Factorial Gaussian Topic Embeddings guided gamma belief network (Gaussian SawETM). 
% A novel deep NTM named Sawtooth Factorial Gaussian Topic Embeddings  guided gamma belief network (SawETM),  is proposed to infer multi-layer document representations and discover topic hierarchies in both embedding space and vocabulary space.
\item To incorporate the structured prior knowledge from the real world, we propose TopicNet, a novel knowledge-based hierarchical topic model based on Gaussian SawETM.
%To the best of our knowledge, TopicNet is the first knowledge-graph guided unsupervised hierarchical probabilistic topic model.
%, which named TopicNet. 
% To move beyond the independence assumption between topics of the two adjacent layers in most topic models, SawETM extends the deep LDA by developing the sawtooth connection in topic embedding space, resulting in more coherent and diverse topic hierarchies.
\item In addition to detailed theoretical analysis, we conduct extensive experiments to verify the effectiveness of the above two models. One of the most appealing properties of TopicNet is its interpretability. We conduct extensive qualitative evaluation on the quality of the topics discovered by TopicNet, including word embedding, topic embedding, and topic hierarchies. 
% \item We propose a hierarchical dependency decoder based on embedding space which higher layer topic can capture lower layer similarly relationships, which help higher layer learn expressive representation.
% \item Aiming at building a very deep generative model, a skip-connected encoder is proposed to approximate the posterior of the latent representation of a document.
% \item we represent a 
\end{itemize}

\section{Gaussian SawETM}\label{sec:Gaussian SawETM}
%Sawtooth Factorial Topic Gaussian Embeddings Guided Gamma Belief Network} \label{sec:Gaussian SawETM}
% As discussed in Section~\ref{introduction}, compared with point embedding, Gaussian embedding provides many interesting advantages, including better capturing uncertainty about a topic embedding and its relationships, expressing asymmetries more naturally than dot product or cosine similarity, and enabling more expressive parameterization of decision boundaries  \cite{das2015gaussian}. So 
In this section, we elaborate on the construction and inference of Gaussian SawETM,
%Sawtooth Factorial Topic Gaussian Embeddings Guided Gamma Belief Network (Gaussian SawETM), 
a deep hierarchical topic model that represents words and topics as Gaussian-distributed embeddings \cite{vilnis2014word}. 
%and then projects the expected likelihood kernel  \cite{jebara2004probability}  of topic Gaussian embeddings at adjacent layers into simplex and acted as topics in GBN  \cite{zhou2015poisson, zhou2016augmentable, duan2021sawtooth}.
\subsection{Symmetric similarity: expected likelihood kernel}\label{subsec2_1}
% and the dot product between two means of independent Gaussian is a perfectly valid measure of similarity (it is the expected dot product). a popular is the dot product between 
% two point embedding.
The distance measure plays a key role in quantifying the similarities between embeddings. A simple choice is taking the inner product between the means of two Gaussians, which, however, %discards the covariances and 
does not  take the advantage of the semantic uncertainties brought by Gaussian-distributed embeddings.
% Since we adopt the Gaussian-distributed embeddings, a preferred symmetric measure is the dot product between the means of two Gaussians (Note that it equals the expected dot product). However, such a measure does not utilize the covariances and would not allow us to enjoy the benefit of semantic uncertainties brought by probability distributions. Therefore, as a logical next choice,
Here we employ the expected likelihood kernel \cite{jebara2004probability, vilnis2014word, athiwaratkun2018hierarchical} as our similarity measure, which is defined as
% \begin{align}
\begin{equation} \label{EL}
\begin{split}
\mbox{E} ^{\text{(s)}}\left( {{\bm{\alpha} _i},{\bm{\alpha} _j}} \right) = \int_{\xv \in {\mathbb{R}^n}} {N\left( {\xv;{\muv _i},\mathop{\Sigmamat} \nolimits_i} \right)N\left( {\xv;{\muv _j},\mathop{\Sigmamat} \nolimits_j} \right)} d\xv = N\left( {\mathbf{0};{\muv _i} - {\muv _j},\mathop{\Sigmamat} \nolimits_i + \mathop{\Sigmamat}\nolimits_j} \right) ,
\end{split}
\end{equation} where $\bm{\alpha} _i$ is a Gaussian distribution with mean ${\muv _i}$ and diagonal covariance matrix $\mathop{\Sigmamat} \nolimits_i$. As a symmetric similarity function, this kernel considers the impact of semantic uncertainties brought by covariances.  
%Note that, in the topic models, the means of Gaussian distribution can capture the semantics of word or topics, while the variances represent the uncertainties (the level of abstraction). 

% the dot product between two means of independent Gaussian is used in  \cite{duan2021sawtooth}. 

% For the topic models, learning the underlying structure from raw data relies on the co-occurring relationship between words, and the co-occurring relationship is symmetric. 

% and 

%  While the dot product between two means of independent Gaussian is a perfectly valid measure of similarity (it is the expected dot product), it does not incorporate the covariances and would not enable us to gain any benefit from our probabilistic model. As discussed in  \cite{das2015gaussian}, the most logical choice for a symmetric similarity function would be to take the inner product between the distributions themselves. Recall that for two (well-behaved) functions , a standard choice of inner product is 

%Since we aim to discriminatively train the weights of the energy function, and it is always positive, we work not with this quantity directly, but with its logarithm. 

\subsection{Document decoder with sawtooth factorization and Gaussian embeddings}
%: Sawtooth factorial Gaussian \lowercase Topic Embeddings Guided Gamma Belief Network}
\label{subsec2_2}
The generative network (also known as decoder) is one of the core components of topic models. As discussed in Section~\ref{sec:introduction}, to build dependency between topics at two adjacent layers and learn a deep topic hierarchy, we draw experience from the Sawtooth Connection (SC) in SawETM. We 
develop a stochastic decoder by introducing Gaussian-distributed embeddings to better represent the topics in %Gaussian 
SawETM. Formally, the generative model with $T$ latent layers can be expressed as
\begin{equation} \label{Gaussian-SawETM} \small
\begin{split}
& \bm{x}_j^{(1)} \sim \mbox{Pois}({\bm{\Phi} ^{(1)}}\bm{\theta} _j^{(1)}),
\left\{ {\bm{\theta} _j^{(t)}\sim \mbox{Gam}({\bm{\Phi} ^{(t + 1)}}\bm{\theta} _j^{(t + 1)},1/c_j^{(t + 1)})} \right\}_{t = 1}^{T - 1},
\bm{\theta} _j^{(T)} \sim \mbox{Gam}(\bm{\gamma} ,1/c_j^{(T + 1)}) ,  \\
& \left\{ \bm{\Phi} _k^{( t )} = \mbox{Softmax} \left( \log \left( \mbox{E} ^{\mbox{(s)}} ( \bm{\alpha} ^{({t - 1})},\bm{\alpha} _k^{( t ) )} \right) \right) \right\}_{t = 1}^T , \left\{ {\bm{\alpha} _k^{(t)} \sim \bm{N} ( {\xv;\bm{\mu} _k^{ (t)}, \mathop{\Sigmamat} \nolimits_k^{(t)}} )} \right\}_{t = 0}^T ,
\end{split}
\end{equation}
In the above formula, $\xv_j\in\mathbb{Z}^{V}$ denotes the word count vector of the $j^{th}$ document, which is factorized as the product of the factor loading matrix $\bm{\Phi}^{(1)}\in\mathbb{R}_+ ^{V\times K_{1}}$ and gamma distributed factor score $\bm{\theta}_j^{(1)}$ under the Poisson likelihood; and to obtain a multi-layer document representation, the hidden units $\bm{\theta}_j^{(t)}\in\mathbb{R}^{K_t}_+$ of the $t^{th}$ layer are further factorized into the product of the factor loading $\bm{\Phi}^{(t+1)}\in\mathbb{R}_+^{K_{t} \times K_{t+1}}$ and hidden units of the next layer. The top-layer hidden units $\bm{\theta}_j^{(T)}$ are sampled from a prior distribution. In addition, the $k^{th}$ topic $\bm{\alpha}^{(t)}_k$ at layer $t$ follows a Gaussian distribution with mean $\bm{\mu}_k^{\left(t\right)}$ and covariance $\mathop{\Sigmamat}\nolimits_k^{\left(t\right)}$, where the mean vector describes a semantic concept and the covariance matrix reflects its level of abstraction. Note that in the bottom layer $\bm{\alpha}^{(0)}$ represents the distributed embeddings of words. And $\bm{\Phi}^{(t)}_k$,  used to capture the relationships between topics at two adjacent layers, is calculated based on the topic representations in the shared embedding space, instead of being sampled from a Dirichlet distribution. In particular, in the equation ${\mbox{E}^{\text{(s)}}}(\cdot)$ refers to the symmetric similarity function defined in Eq.~\eqref{EL}.  
% Compared with SawETM, Gaussian SawETM can capture
% The Probability Product Kernel of topic Gaussian embeddings at adjacent layers ($\bm{\alpha^{(l)}_k}$ and $\bm{\alpha^{(l-1)}_k}$ ) is projected into simplex and result in the factor loading matrix $\bm{\Phi}^{(l)}_k$. $K_l$ denotes the topic numbers at layer $l$. 

% Note that, $\bm{\Phi}^{(l)}_k$ captures the relationship between topics of two adjacent layers and is calculated by the SC technique \cite{zhang2018whai}.  In detail, SC first calculates the semantic similarities between topics of two adjacent layers by the Expected Likelihood of their distribution, and then applies normalization to make sure the sum of each column of $\bm{\Phi^{(l)}}$ is equal to one. Specially, each words and topics is represented as Gaussian embedding. To capture the dependency relation between different layers and map the words and topics at a same space, Gaussian SawETM utilizes the sawtooth connection (SC) to couple hierarchical topics across all layers  \cite{duan2021sawtooth}.

% Note that in SC, the factor loading at layer $l$ is the factor score at layer $l-1$, which distinguishes SawETM from other NTMs. $\bm{\alpha^{(l)}_k} \in \mathbb{R}^{D}$

\subsection{Document encoder: Weibull upward and downward encoder networks}

% we fellow  \cite{zhang2018whai, duan2021sawtooth} use Weibull upward and downward encoder to inder multilayer document representation. Specially, to circumvent the challenging optimization of the gamma distributed conditional posterior of $\bm{\theta^{(l)}}$ shown in \ref{Gaussian-SawETM} , and move beyond deterministic encoder,

As presented in Section \ref{subsec2_2}, instead of using Gaussian latent variables like most of neural topic models~\cite{srivastava2017autoencoding}, our generative model employs the gamma distributed latent variables that are more suitable for modeling sparse and non-negative document representations. 
While in sampling based inference, the gamma distribution is commonly used to represent the conditional posterior of these latent variables, 
%By rationality, a gamma distribution based inference network (encoder) is also needed to approximate the conditional posteriors of gamma latent variables. However, 
the difficulty of reparameterizing a gamma distributed random variable makes it difficult to apply it to an inference network \cite{Fan2020bayesian,Zhang2021bayesian}. 
%hard to efficiently update the network parameters via stochastic gradient descent. 
For this reason, we utilize a Weibull upward-downward variational encoder inspired by the work in \citet{zhang2018whai, zhang2020deep}. We let
\begin{equation} \label{eq_Weibull}
\small
\begin{split} 
q(\bm{\theta} _j^{(t)}\given - ) = \mbox{Weibull}(\bm{k}_j^{(t)},\bm{\lambda} _j^{(t)}),
\end{split}
\end{equation}
where, the parameters $\bm{k}_j^{(t)},\bm{\lambda} _j^{(t)} \in {\mathbb{R}^{{K_t}}}$ are deterministically transformed from both the observed document features and the information from the stochastic up-down path $\bm{\theta} _j^{(t+1)}$. In detail, the inference network can be expressed as
\begin{equation} \label{eq_pgcn} \small
\begin{split}
& \hv_j^{'(0)} = \mbox{ReLU}(\Wmat_1^{(0)}\xv_j + \bv_1^{(0)}))  \quad \hv_j^{'(t)}= \hv_j^{'(t - 1)} + \mbox{ReLU}(\Wmat_1^{(t)}\hv_j^{'(t - 1)} + \bv_1^{(t)}), \quad t=1, \cdots, T, \\ 
& \hv_j^{(t)} = \hv_j^{'(t)} \oplus \bm{\Phi}^{(t+1)} \bm{\theta}_j^{(t+1)}, \quad\quad  t=0, \cdots, T-1, \quad\quad \hv_j^{(T)} = \hv_j^{'(T)} ,\quad  \\
&\kv_j^{(t)}= \mbox{Softplus}(\Wmat_2^{(t)}\hv_j^{(t)} + \bv_2^{(t)}), \quad \lambdav _j^{(t)}= \mbox{Softplus}(\Wmat_3^{(t)}\hv_j^{(t)} + \bv_3^{(t)}), \quad t=0, \cdots, T,
\end{split}
\end{equation}
where $\{ \bv_i^{(t)} \}_{i=1,t=1}^{3,T} \in \mathbb{R}^{K_t} $, $\{ \Wmat_i^{(t)} \}_{i=1,t=1}^{3,T} \in \mathbb{R}^{K_t \times K_{t-1}} $, and $\{ \hv_j^{(t)} \}_{j=1,t=1}^{N,T} \in \mathbb{R}^{K_t} $, $\oplus$ denotes the concatenation in feature dimension, $\mbox{ReLU}(\cdot) = \mbox{max}(0, \cdot) $ is the nonlinear activation function, 
and $\mbox{Softplus}(\cdot) $ applies $\mbox{ln}[1 + \mbox{exp}(\cdot)] $ to each element to ensure positive shape and scale parameters of the Weibull distribution. To reduce the risk of posterior collapse, a skip-connected deterministic upward path \cite{duan2021sawtooth} is used to obtain the hidden representations
$\{ \hv_j^{'(t)} \}_{j=1,t=1}^{N,T} $ of the input $\bm{x}_j$. 
\subsection{Inference and estimation} \label{Inference and Estimation}
Similar to variational auto-encoders (VAEs) \cite{kingma2013auto, rezende2014stochastic}, the training objective of Gaussian SawETM is to maximize the Evidence Lower Bound (ELBO):
\begin{equation} \label{ELBO}
\small
\begin{split}
L _{\text{ELBO}} = \sum_{j = 1}^N {\mathbb{E}}\left[ {\ln p({\bm{x}_j}|{\bm{\Phi} ^{(1)}},\bm{\theta} _j^{(1)})} \right]  - \sum_{j = 1}^N {\sum_{t = 1}^T {{\mathbb{E}}\left[ {\ln \frac{{q(\bm{\theta} _j^{(t)}| - )}}{{p(\bm{\theta} _j^{(t)}|{\bm{\Phi} ^{(t + 1)}},\bm{\theta} _j^{(t + 1)})}}} \right]} },
\end{split}
\end{equation}
where $q(\bm{\theta} _j^{(t)}| - )$ is the variational Weibull distribution in Eq.~\eqref{eq_Weibull}, and  $p(\bm{\theta}^{(l)}_j)$ is the prior gamma distribution in Eq.~\eqref{Gaussian-SawETM}. The first term is the expected log-likelihood or reconstruction error, while the second term is the Kullback--Leibler (KL) divergence that constrains $q(\bm{\theta}^{(l)}_j)$ to be close to its prior $p(\bm{\theta}^{(l)}_j)$ in the generative model. The analytic KL expression and simple reparameterization of the Weibull distribution make it simple to estimate the gradient of the ELBO, with respect to $\{\bm{\mu} _k^{\left( t \right)}\}^{T}_{t=0}$ and  $\{ \mathop{\Sigmamat} \nolimits_k^{\left( t \right)}\}^{T}_{t=0}$ of Gaussian embeddings and other parameters in the inference network. 
\section{TopicNet: Semantic graph-guided topic discovery}  
In Section \ref{sec:Gaussian SawETM}, a novel hierarchical topic model called Gaussian SawETM is proposed to discover meaningful topics organized into a hierarchy. Below we describe our solution to incorporate prior belief, $e.g.$, knowledge graph, into Gaussian SawETM to guide the learning of the topic hierarchy. 
% Gaussian distrubution can express asymmetries more naturally than dot product or cosine similarity, which is necessary to represent inclusion or entailment. And as a neural topic model，Gaussian SawETM is easily guided by meta information with the regular term loss. Benefit from this two good properties, we can naturally capture hierarchical structure between topics, and we introduce TopicNet in this section, which aim to integrate prior knowledge structure between concepts to deep topic model.  To combine Hierarchical topic model with WordNet, we first build a hierarchical tree form the concepts in ImageNet. Besides, due to different domain of datasets, we build a specific TopicTree from WordNet that is applicable to a specific dataset. And then we introduce how to keep this TopicTree structure in deep topic model, in the semantic embedding space by density order embedding method. 
%Hierarchical topic models is structured as a tree, while WordNet is structured as a directed graph. For example, a ``dog'' is both a type of ``canine''  and a type of ``domestic animal'' which are both synsets in WordNet. 
\begin{figure*}[!t]
 \centering
  \subfigure[]{\includegraphics[width=8mm]{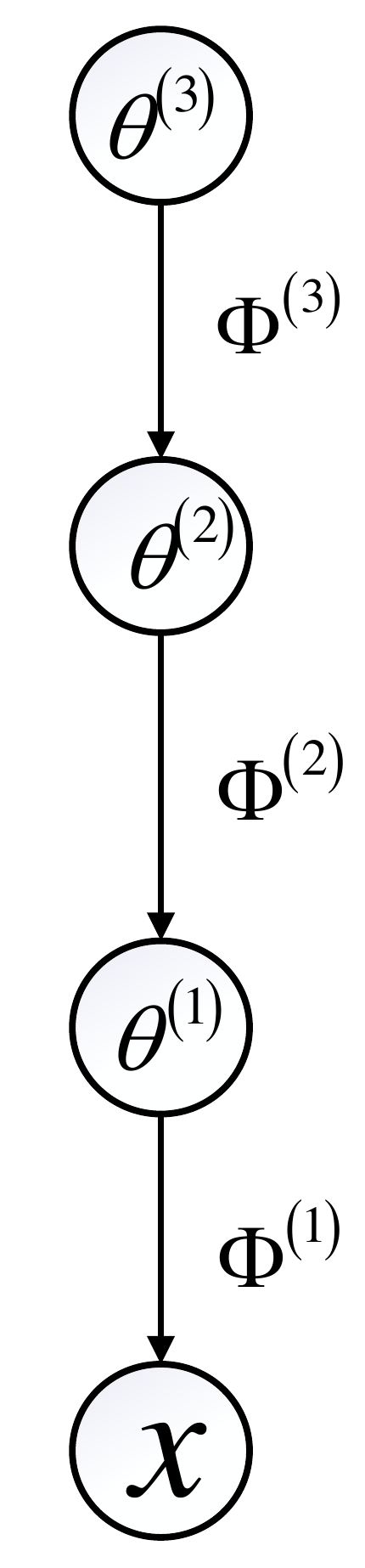}
  \label{fig:generate}}
  \quad \quad
  \subfigure[]{\includegraphics[width=56mm]{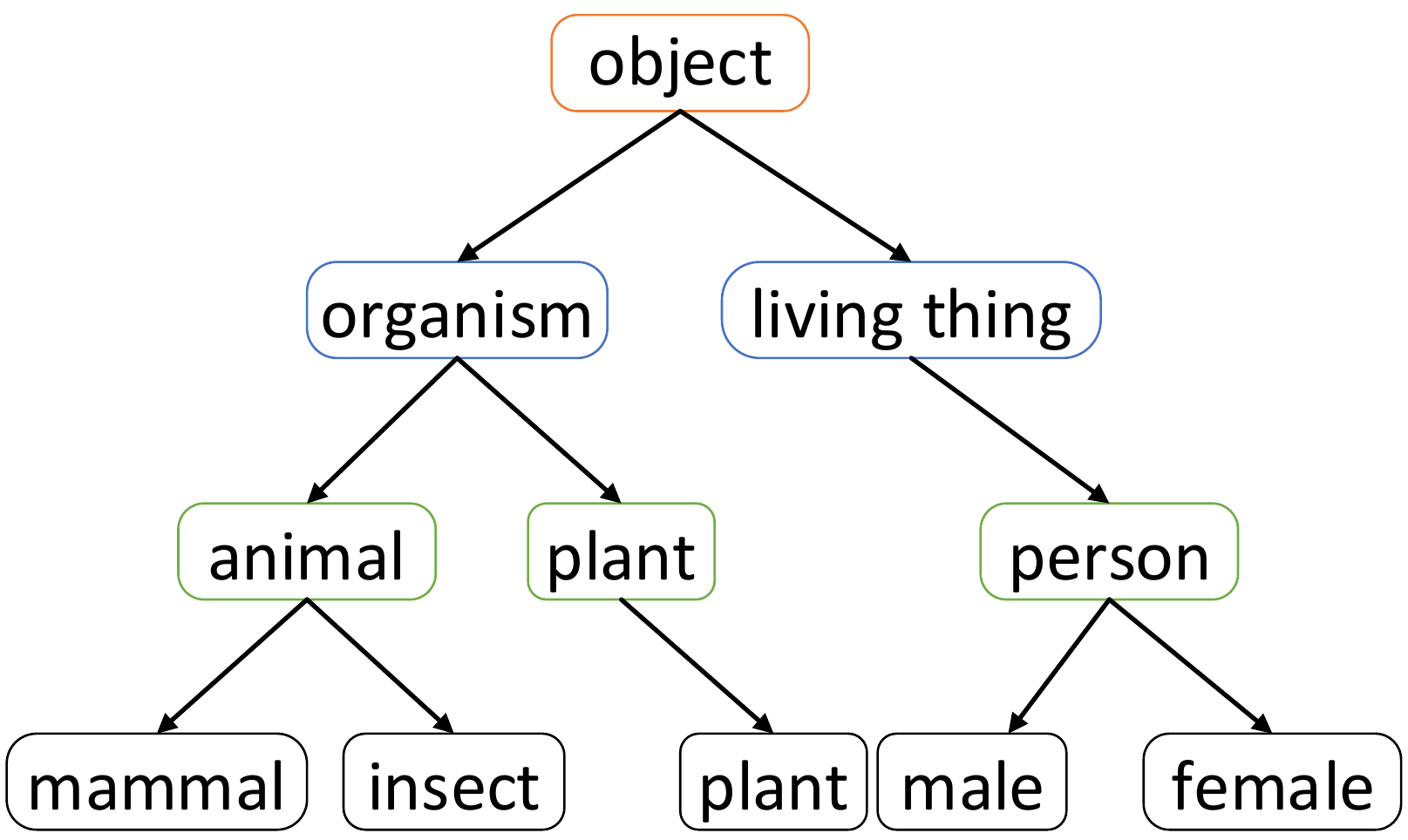}
  \label{fig:wordNet}}
  \quad 
  %\quad \quad \quad
  \subfigure[]{\includegraphics[width=48mm]{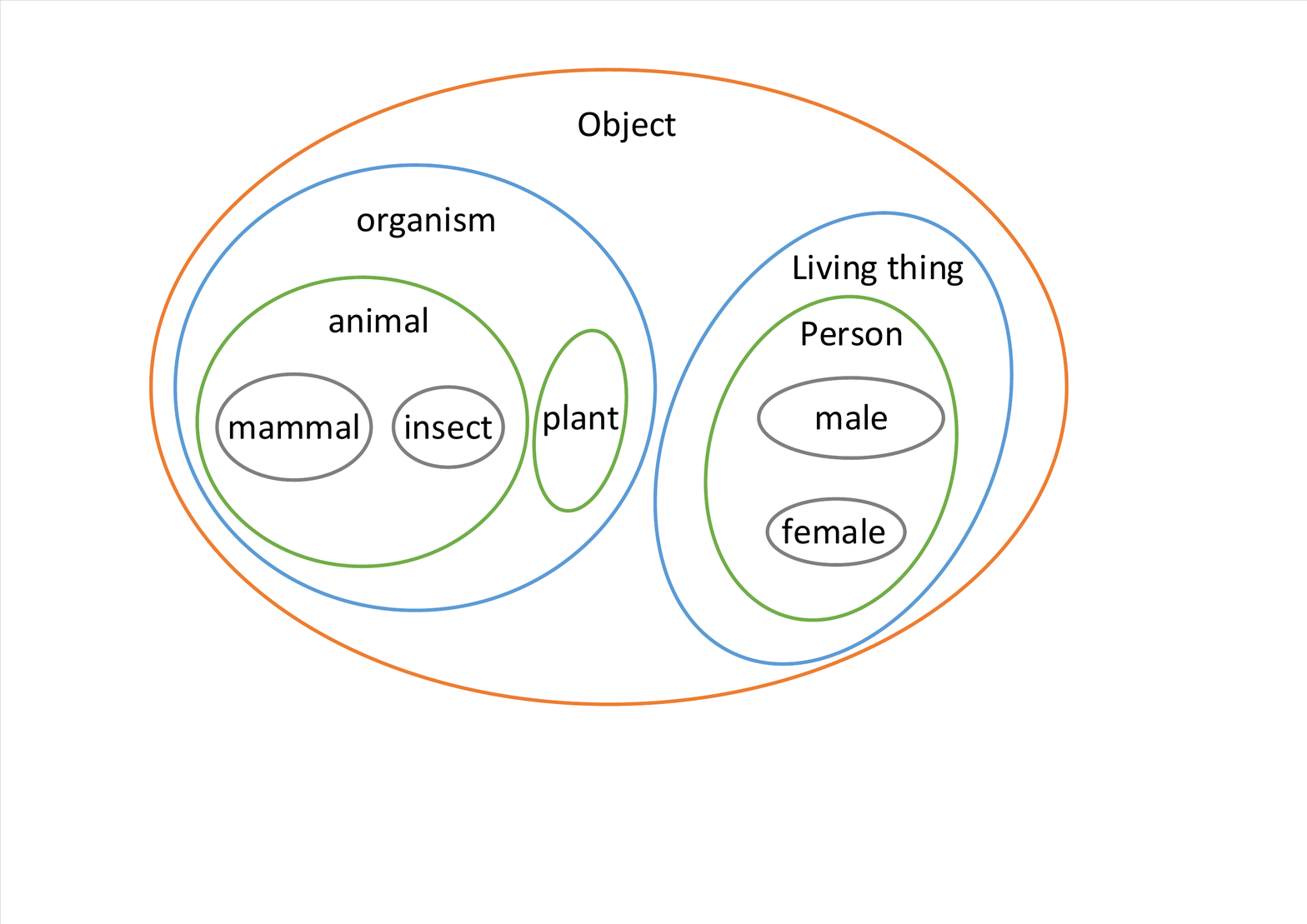}
  \label{fig:wordNet_gaussian_embedding}}
  \caption{\small Overviews of (a) generative model of Gaussian SawETM, (b) TopicTree constructed by WordNet, and (c) Density order embeddings where specific entities correspond to concentrated distributions encapsulated in broader distributions of general entities. }
 \label{model}
\end{figure*} 
 %\vspace{-3mm}
\subsection{TopicTree: Semantic graph constructed by prior knowledge}\label{sec:topictree}  
% \begin{algorithm}[tb]
%   \caption{Building TopicTree}
%   \label{alg:example}
% \begin{algorithmic}
%   \STATE Set mini-batch size $m$ and the number of layer $L$
%   \STATE Initialize the encoder parameters $\Omegav $ and decoder parameters $\bm{\Psi}$;
%   \FOR{$iter = 1,2, \cdot \cdot \cdot $} 
%   \STATE Randomly select a mini-batch of $m$ documents to form a subset ${\rm{X}} = {\left\{ {{\xv_i}} \right\}_{1,m}}$;
%   \STATE Dram random noise $\left\{ {\bm{\varepsilon} _i^l} \right\}_{i = 1,l = 1}^{m,L}$ from uniform distribution;
%     \STATE Calculate  $\nabla {}_{\Omegav , {\bm{\Psi}}}L\left( {\Omegav ,{\bm{\Psi}};{\rm{X,}} \left\{ {{\bm{\varepsilon} _i^l}} \right\}_{i = 1,l = 1}^{m,L}} \right)$ according to Eq.~\eqref{ELBO}, and update encoder parameters $\Omegav $ and decoder parameter $\bm{\Psi}$ jointly ;
%   \ENDFOR
% \end{algorithmic}
% \end{algorithm} 
Finding an appropriate way to encode prior knowledge is especially important for realizing the incorporation of prior belief into topic models. Due to the hierarchical structure of deep topic models, correspondingly, we represent prior knowledge in the form of a hierarchical semantic graph and we call it TopicTree. Particularly, in the TopicTree each node is described by a semantic concept and the concepts of nodes at two adjacent layers follow the entailment relations, which results in a bottom-up process of gradual abstraction, as shown in Fig. \ref{fig:wordNet}. Besides, it is worth mentioning that some generic knowledge graphs such as WordNet \cite{miller1995wordnet} provide us a convenient way to build the TopicTree, since the hypernyms or hyponyms of any given word can be easily found in them.  
\subsection{Partial order: Asymmetric entailment relationship} 
As described in Section \ref{sec:topictree}, 
% Given a TopicTree
we organize prior knowledge into a hierarchical semantic graph so that it can be incorporated into our Gaussian SawETM naturally. Specifically, for a constructed TopicTree with depth $T$, we build a Gaussian SawETM with $T$ latent layers corresponding to it, then we match the topics of each layer in the Gaussian SawETM with the semantic concepts of the corresponding layer in the TopicTree, $i.e.$, each topic has a corresponding semantic concept to describe it. Most significantly, the relationships between semantic concepts in the TopicTree should also be injected
to guide the learning of the topic hierarchy. Since the Gaussian SawETM projects the topics of all layers to a shared embedding space, such a goal can be achieved by imposing constraints on the topic representations in the shared embedding space.
 
Inspired by \citet{vendrov2015order}, the semantic hierarchy (akin to the hypernym relation between words) of the TopicTree can be seen as a special case of the partial order structure. In detail, a partial order over a set of points $X$ is a binary relation $\preceq$ such that for $a,b,c \in X$, the following properties hold: 1) $a \preceq a$ (reflexively); 2) if $a \preceq b$ and $b \preceq a $ then $a=b$ (anti-symmetry); 3) if $a \preceq b$ and $b \preceq c$ then $a \preceq c$ (transitivity). %Thereupon, 
\citet{vendrov2015order} propose learning such asymmetric relationships with order embeddings:  they represent semantic concepts as vectors of non-negative coordinates, and stipulate that smaller coordinates imply higher position in the partial order and correspond to a more general concept. For example, the following expression defines an ordered pair $(x,y)$ of vectors in $\mathbb{R}_+ ^N$:
\begin{equation}\label{eq:order_ebmd}
\small
\begin{split}
\xv \preceq \yv \text{ if and only if} \bigwedge_{i = 1}^N \xv_i \geq \yv_i . 
\end{split}
\end{equation}
Since the order embedding condition defined by Eq. \eqref{eq:order_ebmd} is too restrictive to satisfy by imposing a hard constraint, \citet{vendrov2015order} relax the condition and turn to seek an approximate order-embedding. In particular, for an ordered pair of points, they apply a penalty that is formulated as:
\begin{align}\label{eq:order_penalty}
\Ev(\xv,\yv) = {\|\max (0,\yv-\xv)\|}^2.
\end{align}
Note that this imposes a strong prior on the embeddings, which could encourage the learned relations to satisfy the partial order properties of transitivity and antisymmetry.

Although Gaussian-distributed embeddings are used to represent topics in our Gaussian SawETM, we find that they are also natural at capturing the partial order structure (or the semantic hierarchy of the TopicTree) \cite{das2015gaussian, athiwaratkun2018hierarchical}. For instance, for two topics following the hypernym relation, their Gaussian-distributed embeddings $\bm{\alpha}_i$ and $\bm{\alpha}_j$ form an ordered pair. Assuming that $\bm{\alpha}_i$ represents the specific topic and $\bm{\alpha}_j$ represents the general one, then we have $\bm{\alpha}_i \preceq \bm{\alpha}_j$, meanwhile we expect that $\bm{\alpha}_j$ corresponds to a broad distribution that encompasses $\bm{\alpha}_i$. To meet this expectation, the KL divergence between $\bm{\alpha}_i$ and $\bm{\alpha}_j$ provides us a ready-made tool:
%\begin{small}
% \mz{use resize box rather small font}
\begin{equation} \label{KL}
\small
\begin{split}
\mbox{\Ev}^{\text{(a)}} \left( {{\bm{\alpha} _i},{\bm{\alpha} _j}} \right) = {D_{KL}}\left( {{\Nv_i}\parallel {\Nv_j}} \right) = \int_{\bm{x} \in {\mathbb{R}^n}} {N\left( {\bm{x};{\bm{\mu} _i},{\bm{\Sigma} _i}} \right)\log \left( {{{N\left( {\bm{x};{\bm{\mu} _i},{\bm{\Sigma} _i}} \right)} \mathord{\left/
 {\vphantom {{N\left( {\bm{x};{\bm{\mu} _i},{\bm{\Sigma} _i}} \right)} {N\left( {\bm{x};{\bm{\mu} _j},{\bm{\Sigma} _j}} \right)}}} \right.
 \kern-\nulldelimiterspace} {N\left( {\bm{x};{\bm{\mu} _j},{\bm{\Sigma} _j}} \right)}}} \right)d\bm{x}} \notag \\
  = \frac{1}{2}\left( {{\text{tr}}\left( {\bm{\Sigma} _j^{ - 1}{\bm{\Sigma} _i}} \right) + {{\left( {{\bm{\mu} _i} - {\bm{\mu} _j}} \right)}^{\text{T}}}\bm{\Sigma} _j^{ - 1}\left( {{\bm{\mu} _i} - {\bm{\mu} _j}} \right) - d + \log \left( {{{\det \left( {{\bm{\Sigma} _j}} \right)} \mathord{\left/
 {\vphantom {{\det \left( {{\bm{\Sigma} _j}} \right)} {\det \left( {{\bm{\Sigma} _i}} \right)}}} \right.
 \kern-\nulldelimiterspace} {\det \left( {{\bm{\Sigma} _i}} \right)}}} \right)} \right)
\end{split}
\end{equation}
%\end{small}
However, using $\mbox{E}^{\text{(a)}}({\bm{\alpha}_i},{\bm{\alpha}_j})$ directly as a penalty for violating the partial order is undesirable. Since the KL divergence has a property that $\mbox{E}^{\text{(a)}}({\bm{\alpha}_i},{\bm{\alpha}_j})=0$ if and only if $\bm{\alpha}_i=\bm{\alpha}_j$, which means the penalty is zero only if $\bm{\alpha}_i=\bm{\alpha}_j$, but we expect the penalty should also be zero when $\bm{\alpha}_i \neq \bm{\alpha}_j$ and $\bm{\alpha}_i \preceq \bm{\alpha}_j$. For this reason, following \citet{athiwaratkun2018hierarchical}, we consider using a thresholded divergence as our penalty for order violation:
\begin{equation}\label{eq:distance}
\small
\begin{split}
{d_\gamma }\left( {\bm{\alpha}_i,\bm{\alpha}_j} \right) = \max \left( {0,{\mbox{\Ev} ^{\text{(a)}}({\bm{\alpha}_i},{\bm{\alpha} _j})  - \gamma}} \right).
\end{split}
\end{equation}
So this penalty can be zero if $\bm{\alpha}_i$ is properly encapsulated in $\bm{\alpha}_j$. Nevertheless, note that we no longer have the strict partial order with such a penalty. In other words, the transitivity and anti-symmetry are not guaranteed. Therefore, actually our learned relations also respect an approximate order structure. 

\subsection{Objective: ELBO with a prior regularity}
Essentially, TopicNet is a topic model, so the ELBO still plays a leading role in the training objective. At the same time, a regularization term is also necessary for injecting hierarchical prior knowledge to guide the topic discovery. Therefore, the final objective of TopicNet can be written as:
\begin{align}\label{eq:topicnet}
\Lv _{\text{TopicNet}} = \Lv _{\text{ELBO}} + \beta  \Lv _{\text{prior}},
\end{align}
where $\beta$ is a hyper-parameter used to balance the impact of these two terms. Particularly, for the ELBO term $L _{\text{ELBO}}$, it is the same as defined by Eq.~(\ref{ELBO}). As for the regularization term $L _{\text{prior}}$, we cannot only %consider  %
constrain topics that follow the hypernym relation between two adjacent layers, but the topics that do not follow the hypernym relation should also be considered. This is consistent with the notion of positive and negative sample pairs in contrastive learning. Based on this idea, we use a max-margin loss to encourage the largest penalty between positive sample pairs to be below the smallest penalty between negative sample pairs, which is defined as:
%Based on the idea of contrastive learning, positive and negative examples
%Inspired by the loss function in previous work \cite{vilnis2014word, athiwaratkun2018hierarchical}, with the distance in Eq.~\eqref{eq:distance}, we could use a max-margin loss to capture the hypernym relation between the parent nodes with its children nodes, which encourages positive examples to have zero penalty, and negative examples to have penalty greater than a margin. Specially, for a parent node topic $\mathtt{j ^{(t+1)}}$ at $\mathtt{(t+1) ^{th}}$ layer, the children nodes  of this topic $\mathtt{j ^{(t+1)}}$ at $\mathtt{t ^{th}}$ layer can form a set as $\mathtt{D _j ^{(t+1)}}$, to capture the structure between the parent node with its children node, we defined the loss function as:
\begin{align}\label{eq:rank_loss}
\Lv _{j} ^{(t+1)} = \max(0, m - \max {\left\{ {d_\gamma \left( {i,j} \right)} \right\}_{i \in D}} + \min {\left\{ {d_\gamma \left( {i,j} \right)} \right\}_{i \notin D}}) .
\end{align}
This represents the penalty for the $j^{th}$ topic at ${(t+1)^{th}}$ layer, and $D$ is the set of hyponym topics of topic $j$. For $i \in D$, $(i, j)$ forms a positive pair, for $i \notin D$, $(i, j)$ forms a negative pair. Parameter $m$ is the margin that determines the degree to which the positive and negative pairs are separated. Considering the penalties for all topics in each layer, we obtain the prior regularization term.
%Specially, for a parent node topic $\mathtt{j ^{(t+1)}}$ at $\mathtt{(t+1) ^{th}}$ layer, the children nodes  of this topic $\mathtt{j ^{(t+1)}}$ at $\mathtt{t ^{th}}$ layer can form a set as $\mathtt{D _j ^{(t+1)}}$
%+ \max \left\{ {0,m - {{\left\{ {d\left( {i,j} \right)} \right\}}_{i \notin D}}} \right\}
%where the first term $\max {\left\{ {d_\gamma \left( {i,j} \right)} \right\}_{i \in D}}$ is to \mz{????}, the second term $\min {\left\{ {d_\gamma \left( {i,j} \right)} \right\}_{i \notin D}}$ is to , and the final optimized object is to let the most with the max-margin,.
%And then summing all the loss of parent node in the TopicTree, can result in the prior knowledge supervised loss as:
\begin{align}\label{eq:prior_loss}
\Lv _{\text{prior}} = \sum\nolimits_{t = 0}^{T - 1} {\sum\nolimits_{j = 0}^{{K^{\left( {t + 1} \right)}}} {\Lv_j^{\left( {t + 1} \right)}} }.
\end{align}
% To jointly learn the topic model and density order embeddings, inspired by the loss function of  , we propose a prior knowledge supervised loss function. 
% \begin{align}\label{eq:prior_loss}
% L _{\text{prior} = 0 
% \end{align}
%To balance the impact of ELBO and supervisory losses, we introduce a hyper-parameter $\beta$ into the prior loss, and the TopicNet loss can be expressed as:
With the final objective, TopicNet learns the topics influenced by both the word co-occurrence information and prior semantic knowledge, thus discovering more interpretable topics. Note that, in our experiments, the hyper-parameters are set as $m = 10.0$ and $\beta =1.0$.
% the evidence lowerbound (ELBO) of data log-likelihood $\log p(\Xmat, \Amat)$, which can be expressed as
\subsection{Model Properties} 
In this part, we recap several good properties of TopicNet: Uncertainty, Interpretability, and Flexibility.

\paragraph{Effectiveness of Gaussian-distributed embeddings:} Mapping a topic to a distribution instead of a vector brings many advantages, including: on  one hand, better capturing uncertainty about a representation, where the mean indicates the semantics of a topic and the covariance matrix reflects its level of abstraction; on the other hand, expressing asymmetric relationships more naturally than point vectors, which is beneficial to modeling the partial order structure of semantic hierarchy \cite{vilnis2014word,athiwaratkun2018hierarchical}.
\paragraph{Hierarchical interpretable topics:} In conventional hierarchical topic models, the semantics of a topic is reflected by its distributed words, and we have to summarize a semantically coherent concept manually by post-hoc visualization. In some cases, the topic is too abstract to get a semantically coherent concept, which makes it hard to interpret. Such cases are common in deeper hidden layers. While in TopicNet, we know in advance what semantic concept a certain topic corresponds to, since it is encoded from prior knowledge and used to guide the topic discovery process of the model as a kind of supervision information. Therefore, once the model is trained, the words distributed under each topic are semantically related to the concept that describes it, as shown in Fig.~\ref{fig:hier_structure}. From this perspective, our TopicNet has better interpretability.
% discover interpretable topics by using the structure to guided topic learning. Note that, it is 
\paragraph{Flexibility to incorporate prior belief:} Traditional topic models learn topics only by capturing word co-occurrence information from data, which do not consider the possible prior domain knowledge ($e.g.$, Knowledge Graph like WordNet \cite{fellbaum2010wordnet}). However, TopicNet provides a flexible framework to incorporate the structured prior knowledge into hierarchical topic models. 
\section{Experiments} 
We conducted extensive experiments to verify the effectiveness of Gaussian SawETM and TopicNet. Our code is available at \url{https://github.com/BoChenGroup/TopicNet}.  
\paragraph{Baseline methods and their settings:} We compare Gaussian SawETM with the state-of-the-art topic models: {1. LDA Group}, including: latent Dirichlet allocation(\textbf{LDA}) \cite{blei2003latent}, which is a basic probabilistic topic model; LDA with Products of Experts (\textbf{AVITM})  \cite{srivastava2017autoencoding}, which replaces the mixture model in LDA with a product of experts and uses the AVI for training; Embedding Topic Model (\textbf{ETM}) \cite{dieng2020topic}, a generative model that incorporates word embeddings into traditional topic model. 
{2. DLDA Group}, including: Deep LDA inferred by Gibbs sampling (\textbf{DLDA-Gibbs}) \cite{zhou2015poisson}  and  by TLASGR-MCMC (\textbf{DLDA-TLASGR}) \cite{cong2017deep}. {3. Deep Neural Topic Model (DNTM) Group}, including Weibull Hybrid Autoencoding Inference model (\textbf{WHAI}) \citep{zhang2018whai}, which employs a deep variational encoder to infer hierarchical document representations, and Sawtooth Factorial Embedded Topic Model (\textbf{SawETM}) \citep{duan2021sawtooth}, where topics are modeled as learnable deterministic vectors. For all baselines, we use their official default parameters with best reported settings. 
% As shown in  \citep{cong2017deep}, DLDA-Gibbs and DLDA-TLASGR are state-of-the-art topic modeling methods that clearly outperform a large number of previously propose ones, such as replicated softmax  \cite{hinton2009replicated} and the nested Hierarchical Dirichlet procee  \cite{paisley2014nested}.

 \paragraph{Datasets} Our experiments are conducted on four widely-used benchmark datasets, including 20Newsgroups ({20NG}), Reuters Corpus Volume I ({RCV1}), Wikipedia ({Wiki}), and a subset of the Reuters-21578 dataset ({R8}), varying in scale and document length. 20NG, with a vocabulary of 2$,$000 words, has 20 classes and was split into 11$,$314 training and 7$,$532 test documents. RCV1 consists of 804$,$414 documents with a vocabulary size of 8$,$000. Wiki, with a vocabulary size of 10$,$000, consists of 3 million documents randomly downloaded from Wikipedia using the script provided by \citet{hoffman2010online}. R8, with a vocabulary size of 10$,$000, has 8 classes and was split into 5$,$485 training and 2$,$189 test documents. The summary statistics of these datasets and other implementation details (such as dataset preprocessing and length statistics) are described in the Appendix. 
% size of 2,000 consists of 18,845 documents with 20 categories. Following \cite{yao2019graph}, we build the vocabulary with size 36,534 after removing stopwords and low-frequency words appearing less than 5 times. The average document length is 221. \textbf{2. RCV1}, Reuters Corpus Volume I, consists of 804,414 documents \cite{lewis2004rcv1}. The vocabulary contains 8$,$000 tokens and there are words in one document on average 140. \textbf{3. PG-19} is a using text from books extracted from Project Gutenberg \cite{raecompressive2019}. The dataset contains 28,752 books or 11GB of text. with a vocabulary size of 20,000. We split each book by 1024 tokens, resulting in 613,386 documents. Note that the first two datasets are used for document clusters experiment due to there are ground-truth labels and the last three datasets are used for other experiments.
%  including a subset of the Reuters-21578 dataset(\textbf{R8}), 20 News Groups (\textbf{20NG}) consists of 18,845 documents with a vocabulary size of 2,000, Reuters Corpus Volume 2 (\textbf{RCV2}) and \textbf{Wiki}, which \cite{hoffman2010online}.
%  %In particular, 20NG is small dataset, while RCV2 and Wiki are big datasets.
\begin{table}
    \centering
    \caption{\small Comparison of per-heldout-word perplexity on three different datasets.}
    \label{tab:perplexity}
    \scalebox{0.75}{
    \begin{tabular}{c|c|c|c|c|c|c|c|c|c}
    \toprule
    \textbf{Method} &\textbf{Depth}&\textbf{20NG}&\textbf{RCV1} & \textbf{Wiki} & \textbf{Method} &\textbf{Depth}&\textbf{20NG}&\textbf{RCV1} & \textbf{Wiki}   \\
    \midrule
    %K-means  &68.3 &73.4 &25.9 &55.1 &61.2 &74.7 \\
    %NCut  \cite{shi2000normalized} &69.6 &79.2 &65.9 &80.3 &71.2 &76.5 \\ \hline
    %SVD  &64.3 &74.5 & & & & \\
    LDA \cite{blei2003latent}& 1 & 735 & 942 &1553  &DNTM-WHAI \cite{zhang2018whai}&1& 762& 952&1657 \\
    AVITM \cite{xun2016topic}&1 & 784 & 968& 1703& DNTM-WHAI \cite{zhang2018whai}&5 & 726& 906&1595 \\
    ETM \cite{dieng2020topic}&1 & 742& 951 & 1581& DNTM-WHAI \cite{zhang2018whai}& 15 & 724 &902 &1592 \\
    \midrule
    DLDA-Gibbs \cite{zhou2015poisson}& 1& 702 & - & - &DNTM-SawETM \cite{duan2021sawtooth}& 1& 718& 908&1536 \\
    DLDA-Gibbs \cite{zhou2015poisson}&5 & 678 & - & - &DNTM-SawETM \cite{duan2021sawtooth}& 5&688&873& 1503\\
    DLDA-Gibbs \cite{zhou2015poisson}& 15 & 670 & - & - &DNTM-SawETM \cite{duan2021sawtooth}& 15& 684& 864&1492 \\
    \midrule
    DLDA-TLASGR \cite{cong2017deep} &1&714&912 &1455 &DNTM-GaussSawETM &1 & 714& 904& 1529 \\
    DLDA-TLASGR \cite{cong2017deep} &5&684&877& 1432&DNTM-GaussSawETM &5 & 685&866 & 1477 \\
    DLDA-TLASGR \cite{cong2017deep} &15&673& 842& 1421&DNTM-GaussSawETM &15 & 678& 857&1452 \\
    \bottomrule
    \end{tabular}}
\end{table}
\subsection{Unsupervised learning for document representation} 
\paragraph{Per-heldout-word perplexity:} To measure predictive quality of the proposed model, we calculate the per-heldout-word perplexity (PPL)  \cite{cong2017deep}, on three regular document datasets, $e.g.$, 20NG, RCV1, and Wiki. For deep topic models in the DLDA and DNTM groups, we report the PPL with 3 different stochastic layers $T \in \{1, 5, 15\}$. Specifically, for a 15-layer model, the topic size from bottom to top is set as $\mathbf{K} = [256, 224, 192, 160, 128, 112, 96, 80, 64, 56, 48, 40, 32, 16, 8]$, and the detailed description can be found in the Appendix. For each corpus, we randomly select 80\% of the word tokens from each document to form a training matrix ${\rm{X}}$, holding out the remaining 20\% to form a testing matrix ${\rm{Y}}$. We use ${\rm{X}}$ to train the model and calculate the per-held-word perplexity as
 $$\textstyle \exp \left\{ { - {1 \over {{y_{..}}}}\sum\limits_{v = 1}^V {\sum\limits_{n = 1}^N {{y_{vn}}\ln {{\sum\nolimits_{s = 1}^S {\sum\nolimits_{k = 1}^{{K^1}} {\phi _{vk}^{\left( 1 \right)s}\theta _{kn}^{\left( 1 \right)s}} } } \over {\sum\nolimits_{s = 1}^S {\sum\nolimits_{v = 1}^V {\sum\nolimits_{k = 1}^{{K^1}} {\phi _{vk}^{\left( 1 \right)s}\theta _{kn}^{\left( 1 \right)s}} } } }}} } } \right\},$$ where $S$ is the total number of collected samples and ${y_{ \cdot  \cdot }} = \sum\nolimits_{v = 1}^V {\sum\nolimits_{n = 1}^N {{y_{vn}}} }$.
 As shown in Tab.~\ref{tab:perplexity}, overall, DLDA-Gibbs perform best in terms of PPL, which is not surprising as they can sample from the true posteriors given a sufficiently large number of Gibbs sampling iterations. DLDA-TLASGR is a mini-batch based algorithm that is much more scalable in training than DLDA-Gibbs, at the expense of slightly degraded performance in out-of-sample prediction. Apart from these deep probabilistic topic models, Gaussian SawETM outperforms other neural topic models in all datasets. In particular, compared with Gaussian based single layer neural topic model, WHAI performs better, which can be attributed to two aspects, the first is the effectiveness of Weibull distribution in modeling sparse and non-negative document latent representations, and the second is aggregating the information in both the higher (prior) layer and that upward propagated to the current layer via inference network. Benefiting from the embedding decoder and the Sawtooth Connection that builds the dependency of hierarchical topics, SawETM performs better than WHAI, especially at deeper layers. However, modeling topics with point vectors, SawETM cannot capture the uncertainties of topics. Gaussian SawETM represents the topics with Gaussian-distributed embeddings to capture the uncertainties, and achieves promising improvements over SawETM, especially for modeling complex and big corpora such as Wiki. Note that although the methods in DLDA group get better performance, they require iteratively sampling to infer latent document representations in the test stage, which limits their application \cite{zhang2018whai}. By contrast, Gaussian SawETM can infer latent representations via direct projection, which makes it both scalable to large corpora and fast in out-of-sample prediction.
\begin{table}
    \centering
    \caption{\small Comparison of document classification and clustering performance.}
    \label{tab:graph_cluster}
    \scalebox{0.80}{
    \begin{tabular}{c|c|cc|cc|cc}
    \toprule
    \textbf{Methods}&\textbf{Depth} &\multicolumn{2}{c|}{\textbf{Classification}}
    &\multicolumn{4}{c}{\textbf{Clustering}} \\
    & &\textbf{20NG}&\textbf{R8}    &\multicolumn{2}{c}{\textbf{20NG}} &\multicolumn{2}{c}{\textbf{R8}} \\
    & & & & ACC &NMI &ACC &NMI \\
    \midrule
    %K-means  &68.3 &73.4 &25.9 &55.1 &61.2 &74.7 \\
    %NCut  \cite{shi2000normalized} &69.6 &79.2 &65.9 &80.3 &71.2 &76.5 \\ \hline
    %SVD  &64.3 &74.5 & & & & \\
    LDA \cite{blei2003latent}& 1 &72.6$\pm$0.3 &88.4$\pm$0.5 &46.5$\pm$0.2 &45.1$\pm$0.4 & $51.4\pm$0.4 & 40.5$\pm$0.3 \\
    NVITM \cite{xun2016topic}& 1 &71.5$\pm$0.4 & 87.3$\pm$0.2 &48.3$\pm$0.5 &46.4$\pm$0.3 &52.5$\pm$0.2 &41.2$\pm$0.4 \\
    ETM \cite{dieng2020topic}& 1 &71.9$\pm$0.2& 87.9$\pm$0.4 &49.8$\pm$0.6 &48.4$\pm$0.5 &55.3$\pm$0.4 &42.3$\pm$0.2 \\
    \midrule
    PGBN \cite{zhou2016augmentable}&1&74.2$\pm$0.5 &90.8$\pm$0.5 &46.6$\pm$0.5 &45.4$\pm$0.6 &51.7$\pm$0.4 &40.7$\pm$0.4\\ 
    PGBN \cite{zhou2016augmentable}&4&76.0$\pm$0.3 &91.6$\pm$0.4  &48.2$\pm$0.2 &46.5$\pm$0.5 &54.4$\pm$0.4 &41.2$\pm$0.3 \\ 
    \midrule
    WHAI \cite{zhang2018whai}& 1 &72.8$\pm$0.3 &85.3$\pm$0.2  &49.4$\pm$0.5 &46.5$\pm$0.4 &57.9$\pm$0.5 &42.3$\pm$0.4\\
    WHAI \cite{zhang2018whai}& 4 &73.7$\pm$0.5 &88.4$\pm$0.3 &
    49.5$\pm$0.4 &47.0$\pm$0.3 &60.4$\pm$0.4 &44.1$\pm$0.2 \\ 
    \midrule
    SawETM \cite{duan2021sawtooth}& 1 &70.5$\pm$0.3 &88.6$\pm$0.2 &50.2$\pm$0.2 &48.7$\pm$0.3 &61.2$\pm$0.4 &43.4$\pm$0.4 \\
    SawETM \cite{duan2021sawtooth}& 4 &71.1$\pm$0.2 &90.5$\pm$0.3 &51.2$\pm$0.3 &50.7$\pm$0.2 &63.8$\pm$0.3 &45.9$\pm$0.5 \\
    \midrule
    Gaussian SawETM& 1  &72.8$\pm$0.3 &89.8$\pm$0.4 
    &51.3$\pm$0.4 &50.6$\pm$0.2 &61.5$\pm$0.3 &44.1$\pm$0.2 \\
    Gaussian SawETM& 4   &75.2$\pm$0.4 &91.0$\pm$0.3
    &53.0$\pm$0.3 &51.3$\pm$0.2 &65.2$\pm$0.4 &47.0$\pm$0.3 \\ 
    \midrule
   TopicNet& 1  &71.6$\pm$0.2&88.6$\pm$0.3 & 49.5$\pm$0.3 &46.3$\pm$0.4 & 61.0$\pm$0.2& 43.2$\pm$0.3\\
   TopicNet& 4  &74.3$\pm$0.5&89.5$\pm$0.4 & 50.8$\pm$ 0.3& 47.8$\pm$0.4&64.2$\pm$0.3&46.5$\pm$0.2 \\
  \bottomrule
  \end{tabular}}
\end{table}
\paragraph{Document Classification $\&$ Clustering}
Document-topic distributions can be viewed as unsupervised document representations, to evaluate their quality, we perform document classification and clustering tasks. In detail, we first use the trained topic models to extract the latent representations of the testing documents and then use logistic regression to predict the label and use k-means to predict the clusters. we measure the classification performance by classification accuracy, and clustering performance by accuracy (ACC) and normalized mutual information (NMI), both of which are the higher the better. we test the model performance on 20NG and R8, where the document labels are considered. Different from the perplexity experiments, a larger vocabulary of 20$,$000 words is used for 20News to achieve better performance. The network structure is defined by TopicTree, which is introduced in Section~\ref{sec:topictree}, with the details  deferred to the Appendix. The comparison results are summarized in Tab.~\ref{tab:graph_cluster}. Benefiting from the use of Gaussian distributions on topics, Gaussian SawETM outperforms the other neural topic models on not only  document classification  but also clustering, demonstrating its effectiveness of extracting document latent representations. Note that Gaussian SawETM acquires promising improvements compared to SawETM in the 20NG dataset, showing the potential for modeling more complex data. With the predefined knowledge prior, TopicNet does not perform better, possibly because the introduced prior information does not well align with the end task. We leave the refinement of prior information for specific tasks to future study.
%and we will explore how to build a task-related prior to improve the TopicNet's performance on special tasks. 

% For Table 2 of the original paper, we choose a vocabulary of 20,000 terms, which has 4693 terms that overlap with the vocabulary of the WordNet. Using these 4693 shared terms, we extract a [4693,496,89,11,2] subgraph as the prior WordNet. With not only a larger vocabulary size, but also a wider first topic layer (496 topics instead of 314 topics), it is hence not surprising that a better performance can be achieved for downstream tasks.
% Choosing a vocabulary of 2000 terms (words) for the 20 newsgroups, we compare the 2000 terms to these in the WordNet, which consists of over 70,000 terms, and find 736 terms that are shared between the 20newsgroups and WordNet. Extracting a four-layer subgraph from the WordNet that are rooted at these 736 terms, we obtain a prior WordNet with [736,314,152,52,11] nodes from the bottom to top layers.  
%\mz{do you mean logistic regression/a linear classifier?
% It can be observed that with word embedding and sawtooth connection, SawETM extracts more separable document latent representations  and outperforms the other comparison methods.
% and report the purity and Normalized Mutual Information (NMI) on 20NG, and R8, where the document labels are considered.  With the default training/testing splits of the datasets, we train a model on
% the training documents and infer the topic distributions z on the testing documents.
\subsection{Interpretable topic discovery}
Apart from excelling at document representation learning, another appealing characteristic of TopicNet is that it can discover interpretable hierarchical topics. In this section, we perform the topic discovery experiment with a $7$-layer TopicNet trained on 20NG. The network structure is constructed corresponding to the structure of TopicTree, which is set as $\text{[411, 316, 231, 130, 46, 11, 2]}$. 
% \paragraph{Document Classification $\&$ Clustering}
% We construct the semantic graph from WordNet in all experiments/datasets. 
\paragraph{Prior knowledge and corresponding learned topic:} 
We show six example topics %(out of 314 nodes from the bottom topic layer) 
and their corresponding children nodes in Tab.~\ref{tab:Concept_topic}, which illustrates how a particular concept is being represented in the 20 newsgroups. 
For example, the “military$\_$action” concept, which has “assault”, “defense”, 
and “war” as its children nodes, has become tightly connected to the Waco Siege heavily discussed in this dataset;
the “macromolecule” concept, which has “oil” as its single child node, has been represented by a topic related to vehicle, which is interesting as the 20newsgroups dataset has the following two newsgropus--- “rec.autos’ and “rec.motorcycles”---but not a newsgroup clearly related to biology and chemistry; and the “Atmosphere” concept, which has “Sky” as its single child node, has been represented by a topic related to NASA, reflecting the fact that “sci.space” is one of the 20 news~groups.
%(out of 736 nodes from the bottom word layer) in the Table.~\ref{tab:Concept_topic}

\begin{wrapfigure}{r}{0.5\textwidth}
\begin{center} \vspace{-10mm}
\includegraphics[width=0.45\textwidth]{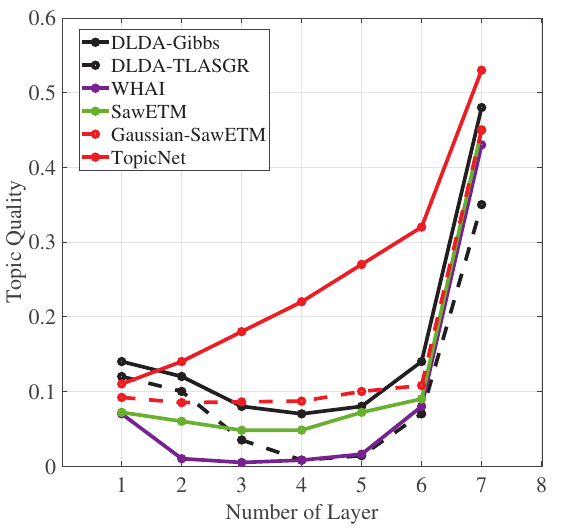} 
\end{center} \vspace{-4mm}
\caption{Topic quality as measured by normalized product of topic coherence and topic diversity (the higher the better) with the varied number of layers. Each layer's topic quality is influenced by its layer size, which is set as $\text{[411, 316, 231, 130, 46, 11, 2]}$ from the bottom layer to the top layer.} 
\label{fig:topic_quality} \vspace{-6mm}
\end{wrapfigure}

\paragraph{Topic quality:} Two metrics are considered here to evaluate the topic quality. The first metric is topic coherence, which is computed by taking the average Normalized Pointwise Mutual Information (NPMI) of the top 10 words of each topic \cite{aletras2013evaluating}. It provides a quantitative measure of the semantic coherence of a topic. The second metric is topic diversity \cite{dieng2020topic}, which suggests the percentage of unique words in the top 25 words of all topics. Diversity close to 1 means more sundry topics. Following \citet{dieng2020topic}, we define topic quality as the product of topic coherence and topic diversity. Note that topic quality is affected by the topic size, so it makes sense to compare different models on the same layer.  As shown in Fig.~\ref{fig:topic_quality}, Gaussian SawETM performs better compared to SawETM, which can be attributed to the Gaussian-distributed embeddings that capture the semantic uncertainties of topics. Incorporating the hierarchical prior knowledge, TopicNet further achieves a higher score than Gaussian SawETM. However, it gets a lower score compared to DLDA-Gibbs in the first two layers, as the layer goes deeper, TopicNet performs better. This is easy to understand since information from data reduces slightly in the shallow layers but severely in the deep layers. 

%And similar to deep VAE, this phenomenon is called mode collapse \cite{maaloe2019biva}. Benefiting from the sawtooth connection(SC) between different layers, the topic information at lower layers can flow to the upper network, which can help the higher layer learn the meaningful topic, and results in better prior learned by SawETM. We can see that SawETM have better performance with a bigger layer size, and get comparable performance with DLDA.
% visitation 
\begin{figure*}[!ht]
\centering
\subfigure[Words Embedding]{
\includegraphics[width=0.28\linewidth]{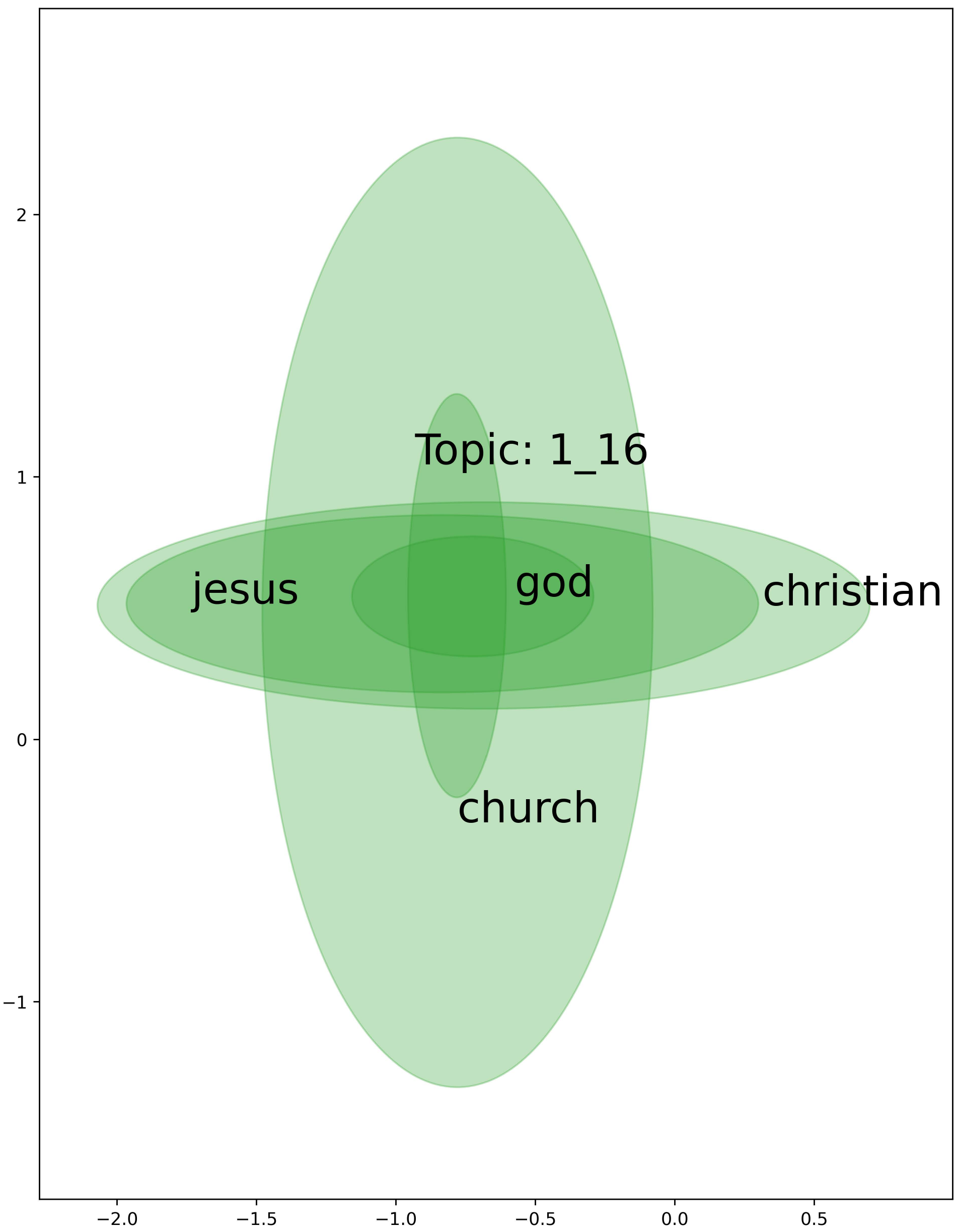}
\label{figure:word_embedding}
}
\quad
\subfigure[Topic Embedding]{
\includegraphics[width=0.60\linewidth]{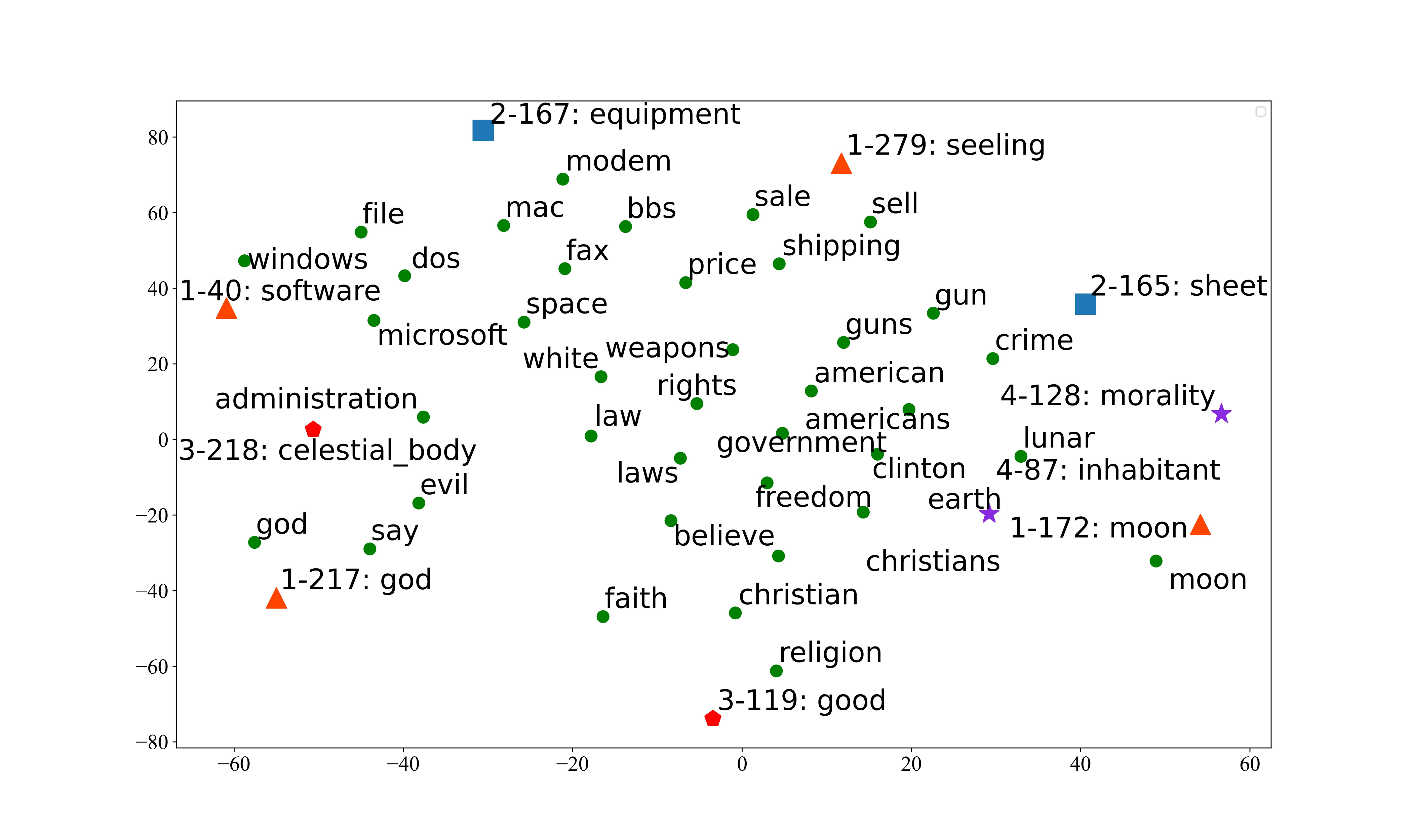}} 
\label{figure:topic_embedding} \vspace{-2mm}
\caption{(a) 2-dim Gaussian embedding in Gaussian SawETM, which we choose the top four words for the topic and (b) T-SNE visualisation of the mean of Gaussian distribution embedding in TopicNet. (The Topic: $\mathtt{t}$-$\mathtt{j: xx}$ denotes the $\mathtt{j ^{th}}$ topic at $\mathtt{t ^{th}}$ layer and $\mathtt{xx}$ represent the concept about the topic.)} 
\label{figure:embedding}
\end{figure*}

% \subsection{Qualitative Analysis}
\begin{table}
    \centering
    \caption{\small Semantic graph and corresponding learned topic on the bottom layer.}
    \label{tab:Concept_topic}
    \scalebox{0.80}{
    \begin{tabular}{c|c|c|c}
    \toprule
    \textbf{Topic}&\textbf{Concept} &\textbf{Children word nodes} &\textbf{Topic words} \\
    \midrule
    %K-means  &68.3 &73.4 &25.9 &55.1 &61.2 &74.7 \\
    %NCut  \cite{shi2000normalized} &69.6 &79.2 &65.9 &80.3 &71.2 &76.5 \\ \hline
    %SVD  &64.3 &74.5 & & & & \\
      11 & military$\_$action & assault defense war & \tabincell{c}{gun fbi guns koresh batf \\ waco assault children compound weapons} \\
    \midrule
      131 & coding$\_$system & \tabincell{c}{address shareware  \\ software unix windows} & \tabincell{c}{windows dos driver microsoft \\ drivers running applications unix using network}\\
    \midrule
     178 & macromolecule & oil & \tabincell{c}{car bike cars engine \\ oil  saturn ford riding miles road}\\
    \midrule
     186 & polity & government & \tabincell{c}{government law rights state \\ police right federal shall crime court laws}\\
    \midrule
     296 & atmosphere & sky & \tabincell{c}{space launch nasa gov \\ satellite earth mission shuttle hst orbit sky}\\
    \midrule
    298 & color & \tabincell{c}{black blue \\ brown color green red} & \tabincell{c}{color green blue red \\ black brown led subject showed lines}\\
    \bottomrule
    \end{tabular}}
\end{table}
\paragraph{Visualisation of embedding space:} First of all, we use a $\text{3}$-layer Gaussian SawETM with $\text{2}$-dim embedding trained on 20NG for the Gaussian-distribution embedding visualization. The top $\text{4}$ words from the $\mathtt{16 ^{th}}$ topic at $\mathtt{1 ^{th}}$ layer are visualized in Fig.~\ref{figure:word_embedding}. As we can see, the words under the same topic are closer in the embedding space, which demonstrates the learned embeddings are semantically similar. Apart from the semantics of words and topics, the shadow range describes the uncertainties and reflects the abstraction levels. Secondly, we visualize the means of Gaussian-distributed embeddings learned by TopicNet with T-SNE \cite{van2008visualizing}. As shown in Fig.~\ref{figure:topic_embedding}, in the shared embedding space, the words with similar semantics are close to each other, meanwhile they all locate near the topic that has similar semantics to them, which means the TopicNet effectively captures the prior semantic information and discovers more interpretable topics. 
\begin{figure*}[!h]
\centering
\includegraphics[width=13cm]{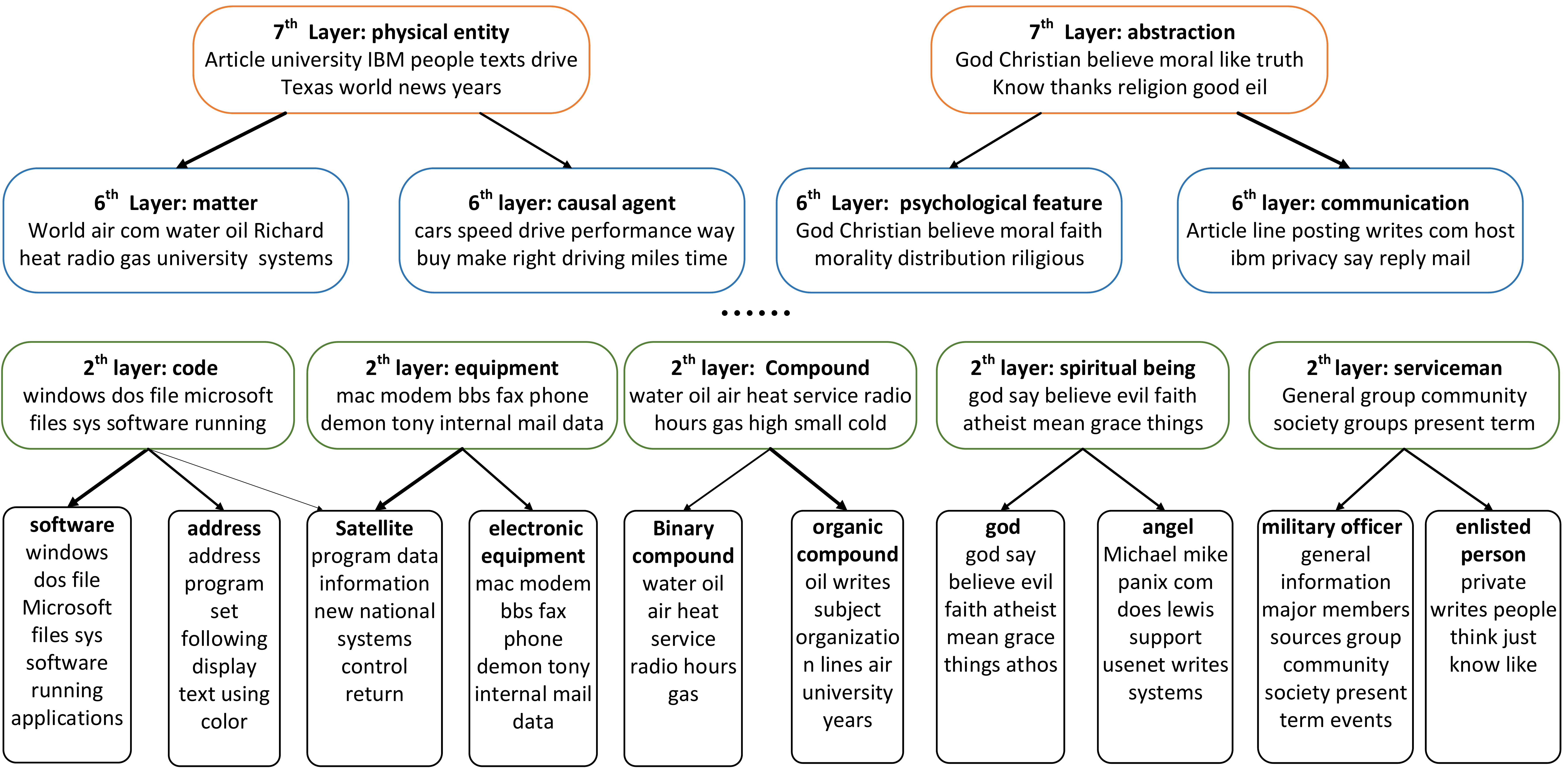} 
\caption{An example of hierarchical topics learned by a 7-layer TopicNet, we only show example topics at the top two layers and  bottom two layers. Boldface indicates the concept of each topic.} \vspace{-5mm}
\label{fig:hier_structure} 
\end{figure*} 
\paragraph{Hierarchical structure of TopicNet:} 
Distinct from many existing topic models that rely on post-hoc visualization to understand topics, we can easily understand the semantic of the whole stochastic network via the predefined TopicTree. Specially, the hierarchical structure of TopicNet is shown in Fig.~\ref{fig:hier_structure}. The semantic of each topic can be represented by the concept in the TopicTree, which is more interpretable compared to traditional topic models. This result further confirms our motivation described in Fig.~\ref{fig:motivation}.\vspace{-2mm}
% we select the top-2 sub-topics of the topic at layer 2. We can see that the sub-topic from a same topic have semantic similarity, and be  The above experiments can prove the SawETM not only learning meaningful word embedding, but also learn meaningful topic embedding.
\section{Conclusion} \vspace{-2mm} 
To equip topic models with the flexibility of incorporating prior knowledge, this paper makes a meaningful attempt and proposes a knowledge-based hierarchical topic model --- TopicNet. This model first explores the use of Gaussian-distributed embeddings to represent topics, and then develops a regularization term that guides the model to discover topics by a predefined semantic graph. Compared with other neural topic models, TopicNet achieves competitive results on standard text analysis benchmark tasks, which illustrate a promising direction for the future development of text analysis models toward better interpretability and practicability. \vspace{-2mm}
% Imposing this inductive bias could help correct data biases, help contextualize the meaning of a concept in the training corpus, and help discover a specific topic hierarchy that would otherwise be shadowed by the dominant topics of the training corpus. A potential data debiasing example: Suppose the training corpus has a biased view of "family" as the composition of "mother," "father," and "son," then we can build a semantic graph, which includes "family" as a parent node and "mother," "father," "son," and "daughter" as its child nodes. Injecting this semantic graph to guide the topic discovery could help correct the data bias of underrepresenting "daughter" as an essential component of the “family'' concept.
% When the semantic graph is not well matched to the main themes of the training corpus, injecting it as a prior will undoublty hurt some downstream tasks, such as document classification and clustering. However, this does not necessarily mean worse performance if our goal is not to boost the performance on these downstream tasks, but to better discover the concepts present in the semantic graph. For example, injecting a semantic graph on "religion'' and contextualizing it under 20Newsgroup will probably lead to bad performance in terms of categorizing the documents into 20 different categories, but may lead to better topic representations of various concepts of religion in 20newsgroup.
\section*{Acknowledgments} \vspace{-2mm}
Bo Chen acknowledges the support of NSFC (61771361), Shaanxi Youth Innovation Team Project, the 111 Project (No. B18039) and  the Program for Oversea Talent by Chinese Central Government. 

\bibliography{neurips_2021}
\bibliographystyle{unsrtnat}
\newpage

\appendix

\section*{Appendix for TopicNet: Semantic Graph-Guided Topic Discovery}
\section{Detailed discussion of our work}
\subsection{Limitations}
This paper proposes a novel knowledge-based hierarchical topic model called TopicNet, which can inject prior knowledge to guide topic discovery. In particular, Gaussian SawETM is first proposed as a common hierarchical topic model, which represents both words and topics by Gaussian-distributed embeddings and builds the dependency between different layers by the Sawtooth Connector module~\cite{duan2021sawtooth}. Experiments show that Gaussian SawETM, which has the ability of capturing semantic uncertainty, gets better performance compared with SawETM, especially for the document classification task. However, Gaussian SawETM requires %a drawback coexisting with the good properties is the
higher memory for training, especially for datasets with big vocabulary size. %which could limit its applicability. 
%Although compared with SawETM, this model only doubles the parameters. 
Due to the high dimension of intermediate variables (e.g. $50000 * 100 * 256 $) in computing the sum of two matrix (e.g. $50000 * 100 * 1  \text{ and }  1 * 100* 256 $),  the gradient of the intermediate variable requires large %a lot of space for 
storage. 
%Different from matrix multiplication which only saves the gradient of two matrix, we solve this problem with this method. 
While the experiments about topic discovery demonstrate that TopicNet %is a fancy model, which 
can express the semantics of each node in the network by concepts, % However, 
as an exploratory model, 
TopicNet guided by a pre-defined graph may not provide better performance for certain specific quantitative tasks. %does not get promising results in the specific tasks, which can be attributed to the pre-defined graph is task-independent. 
%Besides, there is not a quantified experiment about the topic discovery to verify its effectiveness. Overall, we hope our work can give others some inspirations, and we will continue to improve this work and solve already discovered problems.

%  Besides, the training process takes a longer time and bigger memory, which make it not enough attractive.

\subsection{Broader impact}
The proposed TopicNet can be used for text analysis, such as topic discovery and mining hierarchical document representation. Distinct from traditional topic models, TopicNet can express the meaning of each node in the network by pre-defined concepts, which shows better interpretability. With its interpretable latent features, we can further understand the behavior of the model instead of just knowing a result, this can be attractive and important in some special applications. For instance, to recommend articles with specific topics to users, it is necessary to understand the user's interest and incorporate it as a priori into the model to provide a reasonable recommendation. Users could also try to understand why a certain recommendation has been made by the proposed model, which results in more trust. 

% With this interpretability, our model can be applied to the task, which 
% they are able to provide interpretable latent features, one could try to understand why a certain categorization and recommendation has
% been made by the proposed model for a given article, so more appropriate actions can be taken rather than purely trusting the model itself to make the right decisions.

There is significant recent research interest in pre-trained language models, such as BERT \cite{devlin2018bert}, GPT2 \cite{radford2019language}, which are fine-tuned to achieve the state-of-the-art performance in a variety of natural language processing tasks. Despite their promising performance, 
%they are black-box models, which can not be understood by human beings. At present, 
many researchers tend to strongly rely on the numerical performance but pay less attention to their interpretability. We hope our work can motivate machine learners to focus more on the study of understanding the model.

\section{Detailed figure of the proposed model}
\begin{figure*}[!h]
\centering
\includegraphics[width=9cm]{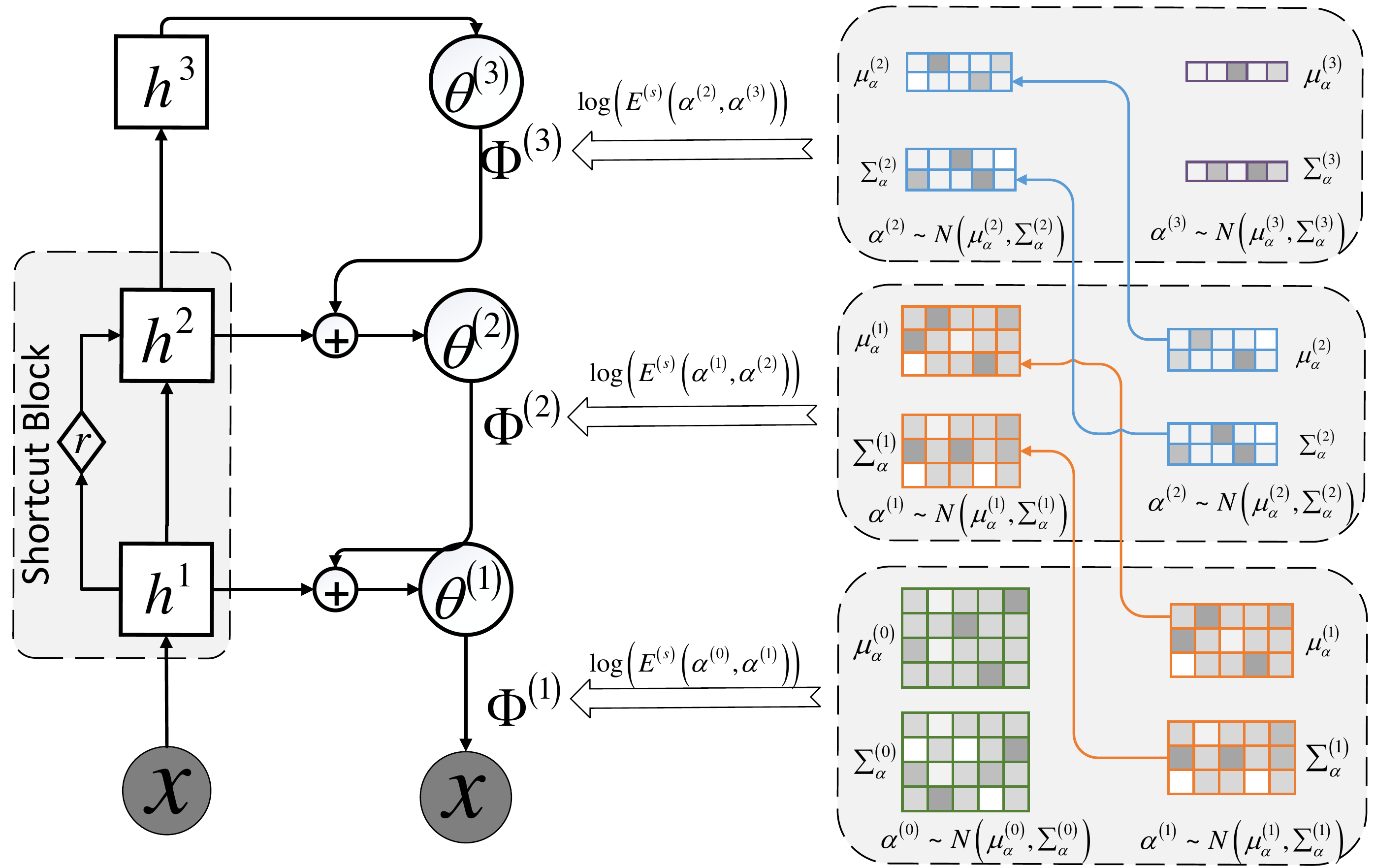} 
\caption{Detailed figure of Gaussian SawETM } \vspace{-5mm}
\label{fig:hier_structure} 
\end{figure*} 

\section{Implementation details}
\paragraph{PPL experiments:} We set the embedding size as 100 and the hidden size as 256. For optimization, the Adam optimizer is utilized here \citep{kingma2014adam} with a learning rate of 0.01. We set the size of minibatch as 200 in 20NG datasets while 2000 in RCV1 and Wiki datasets. All experiments are performed on Nvidia GTX 8000 GPU and coded with PyTorch as attached in the supplement. For the models in the DLDA group and DNTM group, the network structures of 15-layer models are [256, 224, 192, 160, 128, 112, 96, 80, 64, 56, 48, 40, 32, 16, 8], while the network size is set as $\text{256}$ for the methods in LDA group. Due to the network structure of TopicNet is defined by the TopicTree and the main task of TopicNet is topic discovery, so we do not compare TopicNet with traditional topic models on PPL experiments.

\paragraph{Document classification and Clustering experiments:} We set the embedding size as 50 and the hidden size as 256. The 20NG dataset is used with a vocabulary  of size $\text{20,000}$. The network structure is set as [416, 89, 12, 2] for the 20NG dataset and [264, 66, 11, 2] for the R8 dataset. 

\paragraph{Topic quality and visualization experiments:} We use the 20NG dataset with a vocabulary of size $2,000$. The embedding size is set as 50 and the hidden size is set as 1000. The network structure is set as [411, 316, 231, 130, 46, 11, 2]. The other settings are same with previous experiments. 

\section{The details of building TopicTree}
Hierarchical topic models, such as gamma belief network, is structured as a tree where all the leaf nodes of a parent node are on the same floor. 
Due to the complexity of language, the structure of WordNet \cite{miller1995wordnet} is a directed graph but not a tree \cite{redmon2017yolo9000}. So to inject the prior knowledge from WordNet to topic models naturally, we need to first construct a TopicTree that have similar structure with gamma belief network.

\begin{figure*}[!h]\label{fig:example}
\centering
\includegraphics[width=12cm]{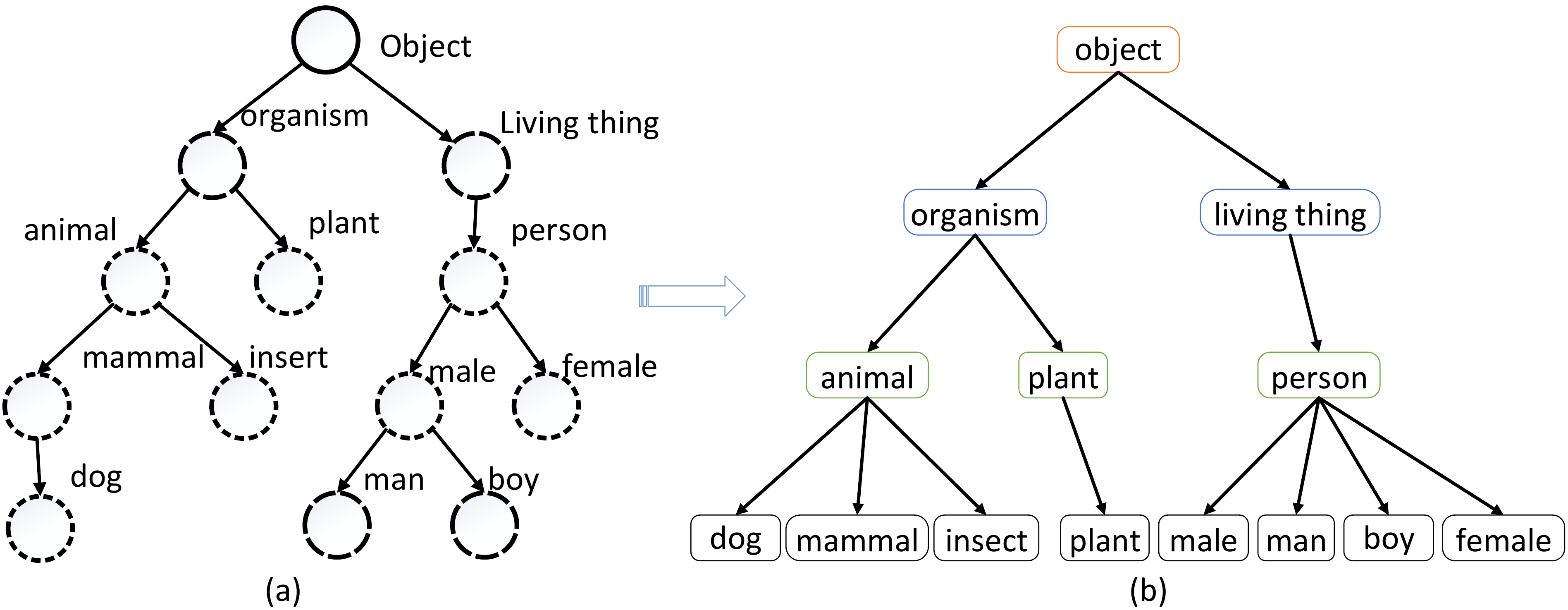}
\caption{An example of (b) $\text{4 layer}$ TopicTree built from (a) WordNet.}
\end{figure*}

\paragraph{Top-down traversal:} For a $T$-layer TopicTree, the original structure for the top $T$ layers in the WordNet is kept. Due to the bottom layer concept is defined as the word layer in the topic model, all the children nodes of the bottom layer node concept are connected to its ancestor node. For example,  as shown in Fig.~\ref{fig:example},  we construct a $\text{4}$-layer TopicTree from the WordNet. At first, the top $t$ layers are kept. Then, the children node ``dog'' of node ``mammal'' is connected to the parent node ``animal.'' The node ``male'' is the same setting. After this process, we can get a TopicTree as shown in Fig.~\ref{fig:example}~(b). Note that, this Tree is built for all the concepts in WordNet, while the vocabulary of the special dataset does not have all the concepts. So to adapt to the domain of the dataset, we need to construct a TopicTree for the special dataset.  

\paragraph{Down-top traversal:} Given the TopicTree as shown in  Fig.~\ref{fig:example}~(b), we need to build a new TopicTree to adapt the domain of a special dataset. Specially, the bottom layer concept is set as word layer,  the intersection of the dataset vocabulary, and then traversing the parent node from the bottom layer to top layer to construct a new TopicTree. 

\section{Other experiment results}
\paragraph{Test time results}
For the perplexity evaluation, the DLDA group requires iteratively sampling to infer latent document representations in the test stage, despite getting better performance. To support this argument, the test time results (average seconds per document) on 20Newsgroup dataset are shown in Tab.~\ref{tab:testing_time}. 
\begin{table}
    \centering
    \caption{\small Comparison of testing time.}
    \label{tab:testing_time}
    \scalebox{0.8}{
    \begin{tabular}{c|c|c}
    \toprule
    \textbf{Methods}&\textbf{Depth} &\textbf{Time(s)} \\
    \midrule
    %K-means  &68.3 &73.4 &25.9 &55.1 &61.2 &74.7 \\
    %NCut  \cite{shi2000normalized} &69.6 &79.2 &65.9 &80.3 &71.2 &76.5 \\ \hline
    %SVD  &64.3 &74.5 & & & & \\
    PGBN \cite{zhou2016augmentable}&1&9.74   \\ 
    PGBN \cite{zhou2016augmentable}&5&11.25   \\ 
    \midrule
    Gaussian SawETM& 1  &0.62 \\ 
    Gaussian SawETM& 5  &0.85 \\ 
    \bottomrule
    \end{tabular}}
\end{table}
\paragraph{Document Classification $\&$ Clustering on 20NG dataset with a vocabulary size of 2000:} We construct the semantic graph from WordNet in all experiments/datasets. Indeed, the quality of the constructed semantic graph has a great impact on the clustering/classification performance. To further illustrate the effectiveness of the proposed model, we build a better-fitted semantic graph for the 20Newsgroups dataset with a vocabulary size of 2000. 
Choosing a vocabulary of 2000 terms (words) for the 20 newsgroups, we compare the 2000 terms to these in the WordNet, which consists of over 70,000 terms, and find 736 terms that are shared between the 20newsgroups and WordNet. Extracting a four-layer subgraph from the WordNet that are rooted at these 736 terms, we obtain a prior WordNet with [736,314,152,52,11] nodes from the bottom to top layers. The  experiment’s results on 20Newsgroup dataset with a vocabulary size of 2000 are summarized in Tab.~\ref{tab:classfication_cluster}. The results show that TopicNet achieves better performance compared with baseline models that do not inject prior knowledge, which confirms our motivation. 
For Tab.~\ref{tab:graph_cluster}, we choose a vocabulary of 20,000 terms, which has 4693 terms that overlap with the vocabulary of the WordNet. Using these 4693 shared terms, we extract a [4693,496,89,11,2] subgraph as the prior WordNet. With not only a larger vocabulary size, but also a wider first topic layer (496 topics instead of 314 topics), it is hence not surprising that a better performance can be achieved for downstream tasks.

\begin{table}
    \centering
    \caption{\small Comparison of document classification and clustering performance.}
    \label{tab:classfication_cluster}
    \scalebox{0.8}{
    \begin{tabular}{c|c|c|cc}
    \toprule
    \textbf{Methods}&\textbf{Depth} &\textbf{Classification}
    &\multicolumn{2}{c}{\textbf{Clustering}}\\
     & & &ACC &NMI \\
    \midrule
    LDA \cite{blei2003latent}& 1 &62.1$\pm$0.4  &37.4$\pm$0.4 &38.1$\pm$0.3  \\
    NVITM \cite{xun2016topic}& 1 &61.3$\pm$0.3  &40.2$\pm$0.3 &41.2$\pm$0.2  \\
    ETM \cite{dieng2020topic}& 1 &61.6$\pm$0.3  &39.3$\pm$0.3 &38.4$\pm$0.4  \\
    \midrule
    PGBN \cite{zhou2016augmentable}&4&63.2$\pm$0.5 &38.4$\pm$0.2 &39.2$\pm$0.3  \\ 
    WHAI \cite{zhang2018whai}& 4 &62.5$\pm$0.4 &38.5$\pm$0.2 &37.2$\pm$0.3  \\ 
    SawETM \cite{duan2021sawtooth}& 4 &62.0$\pm$0.3 &41.6$\pm$0.4 &38.4$\pm$0.2  \\
    \midrule
    Gaussian SawETM& 4   &62.8$\pm$0.3 &42.3$\pm$0.2 &39.0$\pm$0.2 \\ 
   TopicNet& 4  &65.0$\pm$0.3 & 43.2$\pm$0.2& 42.2$\pm$0.4 \\
    \bottomrule
    \end{tabular}}
\end{table}

% \section{Additional result}

% \bibliography{neurips_2021}
% %\bibliographystyle{ieee_fullname}
% \bibliographystyle{unsrt}

\end{document}